\documentclass[10pt,conference]{IEEEtran}
\IEEEoverridecommandlockouts
\usepackage{cite}
\usepackage{amsmath,amssymb,amsfonts,mathrsfs,subfigure,tcolorbox,framed,enumitem,multirow,makecell,threeparttable}
\usepackage[hyphens]{url}
\usepackage{algorithmic}
\usepackage[hidelinks]{hyperref}
\usepackage{graphicx}
\usepackage{textcomp}
\usepackage{xcolor}
\usepackage{balance,caption}
\usepackage{adjustbox,colortbl}

\usepackage[framemethod=tikz]{mdframed}
\newcounter{finding}
\newcommand{\finding}[1]{\refstepcounter{finding}
  \vspace{2.3mm}
 \begin{mdframed}[linecolor=gray,roundcorner=12pt,backgroundcolor=gray!15,linewidth=3pt,innerleftmargin=2pt, leftmargin=0cm,rightmargin=0cm,topline=false,bottomline=false,rightline = false]
  \textbf{Finding \arabic{finding}:} #1
 \end{mdframed}
 \vspace{2.3mm}
}

\usepackage{xspace}
\newcommand{\ours}{FairHOME\xspace}
\newcommand{\ourc}{FairHOME1\xspace}
\newcommand{\ourm}{FairHOME2\xspace}
\newcommand{\oura}{FairHOME3\xspace}
\def\BibTeX{{\rm B\kern-.05em{\sc i\kern-.025em b}\kern-.08em
    T\kern-.1667em\lower.7ex\hbox{E}\kern-.125emX}}

\begin{document}

\title{Diversity Drives Fairness: Ensemble of Higher Order Mutants for Intersectional Fairness of Machine Learning Software}

\author{\IEEEauthorblockN{Zhenpeng Chen\IEEEauthorrefmark{1}, Xinyue Li\IEEEauthorrefmark{2},
Jie M. Zhang\IEEEauthorrefmark{3},
Federica Sarro\IEEEauthorrefmark{4}, 
Yang Liu\IEEEauthorrefmark{1}\IEEEauthorrefmark{5}}
\IEEEauthorblockA{\IEEEauthorrefmark{1}Nanyang Technological University, Singapore}
\IEEEauthorblockA{\IEEEauthorrefmark{2}Peking University, China}
\IEEEauthorblockA{\IEEEauthorrefmark{3}King's College London, United Kingdom}
\IEEEauthorblockA{\IEEEauthorrefmark{4}University College London, United Kingdom}
\IEEEauthorblockA{\IEEEauthorrefmark{5}China-Singapore International Joint Research Institute, China}
\IEEEauthorblockA{zhenpeng.chen@ntu.edu.sg, xinyueli@stu.pku.edu.cn, jie.zhang@kcl.ac.uk, f.sarro@ucl.ac.uk, yangliu@ntu.edu.sg}
\thanks{Corresponding author: Jie M. Zhang (jie.zhang@kcl.ac.uk).}
}

\maketitle

\begin{abstract}
Intersectional fairness is a critical requirement for Machine Learning (ML) software, demanding fairness across subgroups defined by multiple protected attributes. This paper introduces \ours, a novel ensemble approach using higher order mutation of inputs to enhance intersectional fairness of ML software during the inference phase. Inspired by social science theories highlighting the benefits of diversity, \ours generates mutants representing diverse subgroups for each input instance, thus broadening the array of perspectives to foster a fairer decision-making process. Unlike conventional ensemble methods that combine predictions made by different models, \ours combines predictions for the original input and its mutants, all generated by the same ML model, to reach a final decision. Notably, \ours is even applicable to deployed ML software as it bypasses the need for training new models. We extensively evaluate \ours against seven state-of-the-art fairness improvement methods across 24 decision-making tasks using widely adopted metrics. \ours consistently outperforms existing methods across all metrics considered. On average, it enhances intersectional fairness by 47.5\%, surpassing the currently best-performing method by 9.6 percentage points.
\end{abstract}


\section{Introduction}
Machine Learning (ML) software plays a critical role in high-stakes human-related decisions, such as hiring \cite{mahmoud2019performance}, criminal sentencing \cite{donohue2018replacement}, and loan application \cite{wu2019investigations}. In this context, fairness of ML software holds increasing significance, as biases in these systems can perpetuate inequalities and harm historically marginalized groups~\cite{sigsoftChenZSH22,icseZhangH21}. Unfair ML software can lead to ethical, reputational, financial harm, and legal consequences if it violates anti-discrimination laws \cite{Dabs220703277,icseZhangH21}.

Indeed, fairness has long been acknowledged as a foundational requirement for ML software~\cite{sigsoftBrunM18,reHorkoff19,reHabibullahH21,reHabibullahGH23,icseNaharZLK22,baresiREnext,tosemSunCZH24}, aiming at mitigating bias and discrimination tied to protected attributes such as sex, race, and age. From the Software Engineering (SE) perspective, unfairness in ML software can be considered as software `fairness bugs'~\cite{Dabs220710223}. This has led to a growing body of SE studies aimed at improving ML software fairness (also referred to as bias mitigation) \cite{fairsmotepaper,fairmaskpaper,fairwaypaper,sigsoftChenZSH22,icseLiMC0WZX22,sigsoftTaoSHF022,sigsoftNguyenBR23,icseGoharBR23}. 

Software users inherently belong to multiple intersecting identity groups defined by various protected attributes. These intersections yield varied experiences of discrimination among different subgroups. For instance, black women might encounter biases stemming from both sexism and racism. This emphasizes the importance of `intersectional fairness', which measures fairness among subgroups formed by the combination of multiple protected attributes~\cite{ZPICSE24}. 
Notably, intersectional fairness has been encoded in legal regulations~\cite{usequal} and thus highlights the compelling need for software researchers and engineers to consider multiple protected attributes simultaneously \cite{ZPICSE24}. Compared to fairness concerning individual protected attributes, which oversimplifies the complex realities faced by software users, intersectional fairness is considered a more challenging task \cite{SarroRE23}.

Recent SE studies have introduced advanced bias mitigation methods capable of dealing with multiple protected attributes and improving intersectional fairness. Notable examples include FairSMOTE \cite{fairsmotepaper}, which pre-processes training data to mitigate bias across protected attributes and then trains a fairer model; MAAT \cite{sigsoftChenZSH22}, which trains distinct models to optimize fairness for each protected attribute individually and then combines them; and FairMask \cite{fairmaskpaper}, which trains individual extrapolation models based on training data to modify protected attributes in inputs for fairer outcomes.

In this paper, we introduce \ours, a novel ensemble approach using higher order mutation of inputs to improve intersectional fairness of ML software during the inference phase. In the social science domain, it is widely acknowledged that increasing diversity can foster fairness in decision-making~\cite{coleman2017promoting,cimpeanu2023social,kim2017diversity}. Drawing inspiration from this, \ours generates diverse mutated inputs from various subgroups for each given input by applying higher order mutation across multiple protected attributes, thereby enriching the decision-making process with diverse perspectives. This approach is particularly effective in addressing intersectional fairness, which inherently involves numerous subgroups, thus ensuring diversity within the generated input set. Then \ours combines the predictions generated by the ML software for the original input and its mutants to reach the final decision. Unlike conventional ensemble methods \cite{zhou2021ensemble}, including state-of-the-art ones in fairness research \cite{sigsoftChenZSH22,icseGoharBR23}, that combine predictions from different models, \ours combines predictions generated by the same ML model. This unique characteristic enables \ours to bypass the need for creating new models, rendering it even applicable to already deployed ML software.

To evaluate \ours, we conduct a large-scale empirical study, which compares it with seven bias mitigation methods across 24 decision-making tasks using six intersectional fairness metrics. Additionally, given that bias mitigation often decreases ML performance (e.g., accuracy)~\cite{Dabs220703277}, we comprehensively evaluate the fairness-performance trade-off achieved by \ours using an advanced benchmarking tool \cite{sigsoftHortZSH21} with 30 fairness-performance measurements.

The evaluation results demonstrate the effectiveness of \ours in improving intersectional fairness, showcasing its ability to consistently outperform all methods across each metric considered. On average across all metrics, \ours improves intersectional fairness by 47.5\%, marking a 9.6 percentage point improvement over the currently best-performing method. Additionally, \ours achieves this notable fairness improvement with only minimal reductions in ML performance, ranging from 0.1\% to 2.7\% depending on the metric considered. Evaluation using the advanced benchmarking tool~\cite{sigsoftHortZSH21} reveals that \ours surpasses all existing methods in fairness-performance trade-off. 

To summarize, this work offers the following contributions:

\begin{itemize}[leftmargin=*]
\item \textbf{Introduction of \ours}: A novel ensemble approach using higher order mutation of inputs to significantly enhance intersectional fairness in ML software.
\item \textbf{Large-scale empirical study}: An extensive evaluation of \ours against 7 state-of-the-art techniques across 24 decision-making tasks, employing 6 intersectional fairness metrics and 30 fairness-performance measurements.
\item \textbf{Open resources}: Public access to all our data and code~\cite{githublink} to encourage replication and facilitate further research.
\end{itemize}

\section{Preliminaries}\label{prelimi}
\subsection{Fairness in ML Software}\label{fairness_definition}
This paper focuses on fairness of ML classification, the most extensively studied topic in software fairness research \cite{Dabs220710223,DBcorrabs220707068}. In this context, ML software assigns input instances with favorable or unfavorable labels. 

Central to discussions of fairness is the concept of \emph{protected attributes}, which are sensitive characteristics such as sex, race, age, religion, and disability status. The population is divided into groups based on these protected attributes, commonly known as privileged and unprivileged groups. Privileged groups historically enjoy certain advantages, while unprivileged groups face disadvantages or discrimination.
In practice, ML software frequently exhibits bias by more often assigning favorable outcomes to members of privileged groups and unfavorable outcomes to those in unprivileged groups \cite{sigsoftChenZSH22}.

This predicament has spurred researchers to advocate for the principle of group fairness. This principle aims to ensure that decisions made by ML software do not unfairly benefit or harm any specific population group. To achieve this, group fairness requires that the probability of receiving favorable labels remains equal between privileged and unprivileged groups or that the model's performance remains consistent across these groups \cite{icseZhangH21}. Group fairness has garnered substantial attention within the software fairness literature \cite{sigsoftChenZSH22,icseZhangH21,sigsoftHortZSH21,fairsmotepaper,fairmaskpaper,fairwaypaper,icseLiMC0WZX22,sigsoftGalhotraBM17,biswas2020machine} due to its alignment with legal regulations~\cite{ismail2001use}.

\emph{Intersectional fairness} is a critical facet of group fairness, which measures fairness across subgroups defined by the simultaneous presence of multiple protected attributes. It is often used interchangeably with subgroup fairness in the literature \cite{GhoshGR21,ZPICSE24}.
These subgroups are vulnerable to discrimination stemming from the convergence of various unprivileged groups within them, making intersectional fairness essential.

We adopt two intersectional fairness criteria: worst-case intersectional fairness~\cite{ZPICSE24} (also called min-max intersectional fairness~\cite{GhoshGR21})  and average-case intersectional fairness \cite{emnlpSubramanianHBCF21}. Worst-case intersectional fairness quantifies the maximum disparity among subgroups (i.e., the difference between the subgroups with minimum and maximum discrimination), while average-case intersectional fairness calculates the averaged differences between each subgroup and the entire population. 
Additionally, we consider three widely adopted group fairness metrics~\cite{sigsoftChenZSH22,ZPICSE24,icseZhangH21,sigsoftHortZSH21,fairsmotepaper,xinyuedriving}: \emph{SPD} (Statistical Parity Difference), \emph{AOD} (Average Odds Difference), and \emph{EOD} (Equal Opportunity Difference).
Based on the two intersectional fairness criteria and three group fairness metrics, we have six intersectional fairness metrics: WC-SPD, WC-AOD, WC-EOD (for worst-case), and AC-SPD, AC-AOD, AC-EOD (for average-case). 

We assume the presence of $d$ protected attributes denoted as $A_1, A_2, ..., A_d$. Each of these attributes divides the population into various groups. We define a subgroup $sg_{A_1,A_2,...,A_d}$ as the collection of individuals resulting from the intersection of members belonging to groups $g_{A_1}$ through $g_{A_d}$. Formally, this is expressed as $sg_{A_1,A_2,...,A_d} = g_{A_1} \cap g_{A_2}...\cap g_{A_d}$. In this context, building subgroups aims to find all possible \(sg_{A_1, A_2, \ldots, A_d}\) by considering all possible combinations of values for the protected attributes \(A_1, A_2, \ldots, A_d\). For example, consider two protected attributes: sex with groups $g_{sex} \in$ \emph{\{male, female\}} and race with groups $g_{race} \in$ \emph{\{white, non-white\}}. The subgroup set $SG$ includes four subgroups: $SG = $ \emph{\{white male, white female, non-white male, non-white female\}}. We use $Y$ and $\hat{Y}$ to represent the actual decision label and the predicted decision label, respectively, with 1 denoting the favorable label and 0 denoting the unfavorable label. 
Building upon these concepts, we can compute the worst-case intersectional fairness metrics as follows.

\begin{itemize}[leftmargin=*]
\item \textbf{WC-SPD} calculates the maximum difference across subgroups in achieving favorable outcomes:

\begin{equation*}
\scriptsize
\begin{aligned}
\max \limits_{s\in SG} P[\hat{Y} = 1 | sg = s] - \min \limits_{s\in SG} P[\hat{Y} = 1 | sg = s].
\end{aligned}
\end{equation*}

\item \textbf{WC-AOD} calculates the maximum of the average difference in false-positive and true-positive rates across subgroups.

\begin{equation*}
\scriptsize
\begin{aligned}
&\frac{1}{2}[\max \limits_{s\in SG}(P[\hat{Y}=1| sg=s, Y=0] + P[\hat{Y}=1| sg=s, Y=1]) \\ & -  \min \limits_{s\in SG}(P[\hat{Y}=1|sg=s, Y=0] + P[\hat{Y}=1|sg=s, Y=1])].
\end{aligned}
\end{equation*}

\item \textbf{WC-EOD} calculates the maximum difference across subgroups in true-positive rates:

\begin{equation*}
\scriptsize
\begin{aligned}
\max \limits_{s\in SG} P[\hat{Y}=1| sg=s, Y=1]  -  \min \limits_{s\in SG} P[\hat{Y}=1|sg=s, Y=1].
\end{aligned}
\end{equation*}

\end{itemize}

Average-case intersectional fairness metrics calculate the difference between each subgroup and the entire population, and then average these differences \cite{emnlpSubramanianHBCF21}. For instance, AC-SPD computes the favorable rate for each subgroup and the entire population, and then averages the differences between each subgroup's rate and that of the population. Detailed equations are omitted due to the page limit.

\subsection{Related Work}
Fairness improvement, also known as bias mitigation, has garnered growing attention in the research community. From the SE perspective, it aims to address fairness bugs and ensure that software aligns with fairness requirements \cite{Dabs220710223}, thereby emerging as a focal point of discussion in the SE community~\cite{icseZhangH21,sigsoftBiswasR21,sigsoftHortZSH21,icseGoharBR23,Dabs220703277,fairwaypaper,biswas2020machine,icseLiMC0WZX22,fairsmotepaper,fairmaskpaper,sigsoftChenZSH22,sigsoftZhang022,icseGaoZMSCW22,sigsoftNguyenBR23,icseBiswasR23}.

Bias mitigation methods are commonly classified into three types: pre-processing, in-processing, and post-processing \cite{DBcorrabs220707068}. 
Pre-processing methods mitigate bias within training data, preventing its amplification during the training phase and promoting fairness in ML models.
In-processing methods employ optimization strategies to reduce bias during model training.    
Post-processing methods adjust outcomes of ML models to make them fairer.  
Researchers have also been exploring the combination of strategies from different categories. For example, Chakraborty et al. \cite{fairwaypaper} combined a pre-processing strategy called situation testing and an in-processing technique that concurrently optimizes fairness and ML performance.

Bias mitigation for multiple protected attributes is important, as studies have highlighted intersectional fairness issues in practice. For example, Buolamwini and Gebru \cite{fatBuolamwiniG18} conducted an empirical study of commercial gender classification systems and found that darker-skinned females are most frequently misclassified. However, recent research \cite{fairsmotepaper,fairmaskpaper,sigsoftChenZSH22,Dabs220703277,ZPICSE24} has pointed out that existing bias mitigation methods primarily focus on individual protected attributes. 

To alleviate this limitation, SE researchers have proposed techniques that can handle multiple protected attributes simultaneously. Notable examples include FairSMOTE \cite{fairsmotepaper}, MAAT \cite{sigsoftChenZSH22}, and FairMask \cite{fairmaskpaper}. Additionally, there are techniques from the ML community. For instance, Kang et al. \cite{KangXWMT22} framed intersectional fairness improvement as a mutual information minimization problem, using a generic end-to-end algorithmic method to address it. Wang et al. \cite{fatWangRR22} conducted an empirical study of five intersectional fairness improvement methods from the ML community, identifying GRY \cite{icmlKearnsNRW18} as the most effective. GRY improves intersectional fairness by modeling it as a two-player zero-sum game, with the learner as the primal player and the auditor as the dual player.

\ours sets itself apart from these advanced methods by its ability to operate without the creation of new models.
This advantage is particularly relevant for already deployed ML software, where modifying models may not be feasible.
In comparison, FairSMOTE generates instances to balance the distribution of training data across class labels and protected attributes, and trains new models based on the augmented data. MAAT creates fair models through sampling training data and combines them with models optimized for ML performance. Like \ours, it is an ensemble method, but it adheres to the traditional ensemble paradigm~\cite{zhou2021ensemble} that combines multiple models. FairMask modifies inputs by using training data to train extrapolation models that adjust protected attributes in input instances. Differently, \ours operates without the need for  creating extrapolation models. GRY is an in-processing method that needs to retrain models.

Additionally, considerable research efforts have been devoted to the evaluation of bias mitigation methods. For instance, 
Hort et al. \cite{sigsoftHortZSH21} provided a unified perspective by integrating fairness and ML performance, and gauging their interplay. They introduced a benchmarking approach named Fairea, which establishes a unified fairness-performance trade-off baseline for comparing bias mitigation methods. 
Using Fairea, Chen et al. \cite{ZPICSE24} compared existing methods in trade-off between intersectional fairness and ML performance, and found that REW, MAAT, and FairMask can achieve the best results. The three methods are all included as our baseline methods.  
In this paper, we also employ Fairea for fairness-performance trade-off evaluation.

\section{Our approach}
\subsection{\ours: In a Nutshell}\label{nutshell}
From the SE perspective, \ours follows the input debugging paradigm \cite{icseKirschnerSZ20}. 
This paradigm suggests that software issues with processing inputs may not always necessitate modifications to the software itself; instead, adjusting the inputs can also resolve these issues. 

\begin{figure} 
    \centering
\includegraphics[width=1\linewidth]{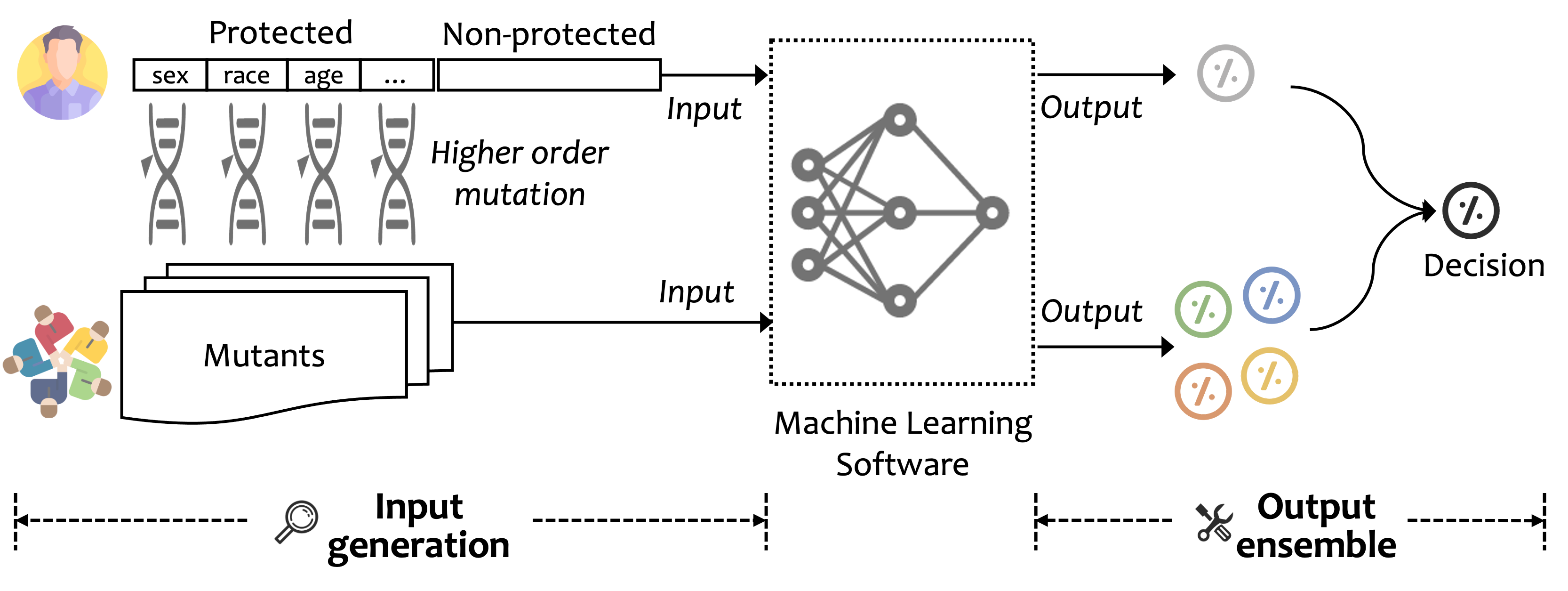}
  \caption{Overview of \ours.}
  \label{fig:approach} 
\end{figure}

Figure \ref{fig:approach} illustrates the overview of \ours. It is inspired by the widely recognized social science insights that increased diversity can foster fairness in decision-making \cite{coleman2017promoting,cimpeanu2023social,kim2017diversity}. Specifically, \ours aims to enhance intersectional fairness by diversifying the inputs used in the decision-making process of the inference phase. This is achieved by generating diverse mutants, which represent different subgroups, for each input instance. \ours employs higher order mutation based on multiple protected attributes to create these mutants.

It then aggregates the decisions of the ML software for the original input and all its mutants to make the final decision.
This marks a departure from traditional ensemble learning \cite{zhou2021ensemble,sigsoftChenZSH22,icseGoharBR23}, which focuses on the ensemble of different models to enhance prediction. Instead, \ours enhances the prediction fairness through the ensemble of outputs from the same ML model.

\subsection{Input Generation}
\ours employs higher order mutation to generate input mutants from an original input $x$ for the ML software $S_{ML}$. This aims to diversify subgroup representation in the decision-making process, thereby addressing intersectional fairness.

To ensure clarity in the subsequent description, we begin by defining the necessary notations. The input for $S_{ML}$ comprises $n$ attributes, collectively represented as $A = \{A_1, A_2,..., A_n\}$, categorized into $d$ protected attributes $\{A_1, A_2,..., A_d\}$ (denoted as $P$) and $n-d$ non-protected attributes $\{A_{d+1}, A_{d+2},..., A_n\}$ (denoted as $N$). Assuming each attribute $A_i$ belongs to a valuation domain $\mathbb{I}_i$, the overall input domain of $S_{ML}$ is $\mathbb{I} = \mathbb{I}_1 \times \mathbb{I}_2 \times ... \times \mathbb{I}_n$.

The detailed higher order mutation process is as follows:

\noindent \textbf{Defining mutation operator}: A mutation operator is defined as a transformation rule that generates a mutant from the original instance \cite{infsofJiaH09}. In our approach, we define the mutation operator $\mu$ to alter the value of each protected attribute $A_i \in P$ within its valuation domain $\mathbb{I}_i$, while keeping the values of non-protected attributes constant. This operation is grounded in the widely adopted causal fairness principle~\cite{Dabs220710223}, which directly modifies protected attributes and assesses the impact of this modification on the prediction \cite{sigsoftGalhotraBM17}.

The mutation operator $\mu$ enables the generation of diverse mutants from a single input instance by applying it across multiple protected attributes.

\noindent \textbf{Generating input mutants}: 
Our input generation process is referred to as higher order mutation in SE \cite{infsofJiaH09}, since it allows for applying the mutation operator to an input instance multiple times, targeting multiple protected attributes. 
Specifically, our higher order mutation entails the use of $\mu$ on protected attributes to produce any mutant $x'$ of $x$ that satisfies two key conditions:

\begin{itemize}
\item There exists at least one protected attribute $a \in P$ for which the value in $x'$ differs from that in $x$ (i.e., $\exists a \in P, x_a \neq {x'}_a$).
\item For each non-protected attribute $q \in N$, the value in $x'$ matches that in $x$ (i.e., $\forall q \in N, x_q = {x'}_q$).
\end{itemize}

Our objective is to exhaustively explore the input domain $\mathbb{I}$ to identify all instances that satisfy these conditions. We define the input domain for protected attributes using all possible value combinations from the training data, ensuring compliance with input constraints. This approach ensures the validity of mutants and keeps the number of possible mutants manageable, limited to combinations present in the training data. This results in a set of mutants $M$ representing diverse subgroups by varying protected attributes. The number of generated mutants is $\prod_{i=1}^{d}|\mathbb{I}_i| - 1$, which represents all possible combinations of protected attribute values, excluding the combination present in the original input instance. For example, with two protected attributes—sex (male, female) and race (white, non-white)—we have four subgroups: white male, white female, non-white male, and non-white female. To produce a fair result for a non-white female, our approach uses inputs for all four subgroups. 

Prior research \cite{icseLiMC0WZX22} has identified correlations between protected attributes and non-protected features, suggesting that mutating non-protected features associated with protected attributes could further enhance fairness. However, this approach requires learning feature correlations. Following the `try-with-simpler' SE practice~\cite{sigsoftFuM17}, we do not adopt this mutation strategy as the default in \ours. Nevertheless, the effectiveness of this alternative strategy, compared to \ours's default, is analyzed in Section \ref{rq4result}.

\subsection{Output Ensemble}\label{outputen}
\ours aggregates outputs of the ML software $S_{ML}$ for the original input and its mutants to produce the final decision, without analyzing output differences. 
Since we focus on ML classification, for each input, $S_{ML}$ produces a probability vector, with each element representing the probability of classification into a category. $S_{ML}$ makes the final decision for each input based on its probability vector.

We introduce three commonly used ensemble strategies~\cite{zhou2021ensemble} to aggregate outputs: majority vote, averaging, and weighted averaging. The details of each strategy are as follows:

\noindent \textbf{Majority vote:} The majority vote strategy selects the decision that receives the most votes as the final decision. Since intersectional fairness typically involves at least two protected attributes, each with at least two values, there are at least three mutants in addition to the original input, making the majority vote strategy applicable. Let $r_0$ represent the decision made by $S_{ML}$ for the original input $x$, and $r_{1}, r_{2}, ..., r_{|M|}$ denote the decisions for each mutant. The strategy assigns a final decision as unfavorable if more than 50\% of the decisions $r_0, r_1, r_2, ..., r_{|M|}$ are unfavorable, and favorable otherwise.

\noindent \textbf{Averaging:} The averaging strategy uses the output probability vectors of the original input and its mutants to determine the final decision. Let $p_0$ represent the probability of the original input being classified as favorable, and $p_{1}, p_{2}, ..., p_{|M|}$ denote such probabilities for each mutant. The strategy computes the mean of $p_0, p_1, p_2, ..., p_{|M|}$. If the final probability is below 50\%, the decision is unfavorable; otherwise, it is favorable.

\noindent \textbf{Weighted averaging:} This strategy uses probability vectors to determine the final decision by calculating the weighted average of the probabilities $p_0, p_1, p_2, ..., p_{|M|}$. The weight $w_i$ for each probability $p_i$ is $|p_i - 50\%|$, which reduces the impact of predictions near the decision boundary, known to be more prone to bias \cite{icseZhangW0D0WDD20,tseZhangWSWDWDD22}. Prediction probabilities close to 50\% indicate a higher risk of bias, so assigning lower weights to these instances minimizes their impact on the ensemble outcome, potentially enhancing fairness. The weighted average is computed as ${W} = \frac{\sum_{i=0}^{|M|} w_i \cdot p_i}{\sum_{i=0}^{|M|} w_i}$. If $W$ is below 50\%, the final decision is unfavorable; otherwise, it is favorable.

Following the `try-with-simpler' SE practice~\cite{sigsoftFuM17}, we use majority vote as the default strategy of \ours due to its simplicity and reliance solely on decision information. Nevertheless, all the strategies are evaluated in Section \ref{rq5result}.

\section{Evaluation Setup}\label{evaluationsetup}

\subsection{Research Questions (RQs)}

\noindent \textbf{RQ1: Can \ours improve intersectional fairness without largely compromising ML performance?} This RQ examines \ours's impact on intersectional fairness and ML performance, considering the trade-off between them~\cite{ase23fairness}.

\noindent \textbf{RQ2: How effectively does \ours improve intersectional fairness compared to existing methods?} This RQ compares the effectiveness of \ours against existing bias mitigation methods in enhancing intersectional fairness. 

\noindent \textbf{RQ3: How does \ours balance intersectional fairness and ML performance compared to existing methods?} This RQ compares \ours with other bias mitigation methods by assessing their fairness-performance trade-off.

\noindent \textbf{RQ4: How effective is \ours when it also mutates features correlated with protected attributes?} This RQ compares the default \ours with its variant that also mutates non-protected features correlated with protected attributes.

\noindent \textbf{RQ5: How do different ensemble strategies affect \ours?} This RQ compares the common ensemble strategies for \ours and evaluates their effectiveness.

\noindent \textbf{RQ6: What is the contribution of different mutants?} This RQ compares the effectiveness of \ours with two approaches: one using only mutants involving a single protected attribute and the other using only mutants involving multiple protected attributes.

\noindent \textbf{RQ7: How does \ours affect group fairness regarding single protected attributes?} This RQ investigates whether \ours negatively impacts group fairness for single protected attributes.

\subsection{Experimental Methodology}

\subsubsection{Step 1. Design of bias mitigation tasks}\label{bmtask}
We use 24 bias mitigation tasks for the study, achieved through a combination of six benchmark datasets and four ML models.

\noindent \textbf{Benchmark datasets:} We select six real-world decision problems using datasets widely adopted in fairness research~\cite{ZPICSE24,sigsoftZhang022,Dabs220703277,sigsoftChenZSH22,fairmaskpaper,fairsmotepaper,icseZhangH21,tosemmengdi}: \emph{Adult}~\cite{adultdata}, \emph{Compas} \cite{compasdata}, \emph{Default} \cite{defaultdata}, \emph{German}~\cite{germandata}, \emph{Mep15} \cite{mep15data}, and \emph{Mep16}~\cite{mep16data}. 

\begin{table*}[!tp]
\small
\centering
\caption{Benchmark datasets.}
\label{dataset_info}
\begin{tabular}{lrlll}
\hline
Name & Size & Protected attributes & Favorable label & Description\\
\hline
Adult \cite{adultdata}  & 48,843 & sex, race & income $>$ 50k & Predicting whether an individual's annual income surpasses \$50K\\
Compas \cite{compasdata} & 7,214 & sex, race, age & no recidivism & Predicting criminal
defendant recidivism\\
Default \cite{defaultdata}& 30,000 & sex, age & default & Predicting whether a customer will
default on payment\\
German \cite{germandata}& 1,000 & sex, age & good credit & Classifying an individual as a good or bad credit risk\\
Mep15 \cite{mep15data}& 15,830 & sex, race & utilizer &  Predicting healthcare utilization using survey data from 2015\\
Mep16 \cite{mep16data} & 15,675 & sex, race & utilizer & Predicting healthcare utilization using survey data from 2016\\
\hline
\end{tabular}
\end{table*}

Table \ref{dataset_info} presents an overview of these datasets. As highlighted in prior research \cite{ZPICSE24}, existing datasets primarily include two protected attributes. The protected attributes for the Adult, Default, Mep15, and Mep16 datasets are specified by Chen et al. \cite{ZPICSE24}, while those for the Compas and German datasets are specified by Zhang et al. \cite{tosemmengdi}.

To our knowledge, our study employs the largest number of datasets in intersectional fairness studies. These datasets are representative for two key reasons: (1) they span diverse domains, such as finance, social, and medical applications, where fairness is crucial, and (2) they cover sex, race, and age, which are demonstrated to be the most commonly considered protected attributes~\cite{corrabs220508809,Dabs220710223}.

\noindent \textbf{ML models:} We use four common types of ML models that are demonstrated to be most widely explored in fairness research~\cite{DBcorrabs220707068}: \emph{LR} (Logistic Regression), \emph{RF} (Random Forest), \emph{SVM} (Support Vector Machine), and \emph{DNN} (Deep Neural Network). 
LR, RF, and SVM have prominent application in fairness-critical decision problems~\cite{ukreport,Dabs220703277}; DNN, due to its significance in modern decision-making scenarios, also garners substantial attention from fairness researchers \cite{Dabs220703277,sigsoftZhang022, icseZhengCD0CJW0C22}. 
For LR, RF, and SVM, we adopt the configurations from recent software fairness papers \cite{Dabs220703277,sigsoftChenZSH22,sigsoftHortZSH21}; for DNN, we adopt a network architecture extensively used for these datasets~\cite{Dabs220703277,sigsoftZhang022, icseZhengCD0CJW0C22}, which features a fully-connected network having five hidden layers with 64, 32, 16, 8, and 4 units, respectively.

For each of the six benchmark datasets, we train the four types of ML models for bias mitigation, resulting in 24 tasks.

\subsubsection{Step 2. Selection of baseline methods}\label{methodselect}
We select seven bias mitigation methods for comparison, covering both widely-adopted and cutting-edge techniques. First, based on a recent survey \cite{DBcorrabs220707068}, we select the three most popular bias mitigation methods: \emph{REW} (Reweighting)~\cite{rewpaper}, \emph{ADV} (Adversarial Debiasing)~\cite{ADVpaper}, and \emph{EOP} (Equalized Odds Post-processing)~\cite{EOpaper}. Second, we include three recently proposed methods in SE, known for their effectiveness in handling multiple protected attributes: \emph{FairSMOTE}~\cite{fairsmotepaper}, \emph{MAAT}~\cite{sigsoftChenZSH22}, and \emph{FairMask}~\cite{fairmaskpaper}. Finally, we consider \emph{GRY}~\cite{icmlKearnsNRW18}, an approach identified as the most effective in a recent empirical study~\cite{fatWangRR22} on intersectional fairness in the ML community.

Our selection represents a diverse range of bias mitigation techniques, including pre-processing, in-processing, and post-processing methods. 

A brief description of each method is provided below:

\begin{itemize}[leftmargin=*]
\item REW~\cite{rewpaper} is a pre-processing method that assigns different weights to training data for each (group, label) combination, thereby ensuring fairness.
\item ADV~\cite{ADVpaper} is an in-processing method that employs adversarial techniques to minimize the influence of protected attributes in predictions, while concurrently maximizing prediction accuracy during model training.
\item EOP~\cite{EOpaper} is a post-processing method that uses a linear program to determine probabilities for altering output labels in order to optimize equalized odds.
\item FairSMOTE~\cite{fairsmotepaper} identifies the largest subgroup in training data, and generates samples for other subgroups to achieve balanced sample counts and equitable favorable rates among subgroups. It also excludes ambiguous training data through situation testing.
\item MAAT~\cite{sigsoftChenZSH22} aims to improve fairness-performance trade-off of ML software. To achieve this, it develops fairness-optimized models for individual protected attributes through training data sampling, and then combines their outputs with a performance-optimized model.
\item FairMask~\cite{fairmaskpaper} uses training data to learn individual extrapolation models that predict each protected attribute using other features. Then it applies these extrapolation models to reassign protected attributes in input data.
\item GRY~\cite{icmlKearnsNRW18} formulates intersectional fairness improvement as a two-player zero-sum game between the learner (primal player) and the auditor (dual player).
\end{itemize}

\subsubsection{Step 3. Analysis of \ours's effect}
This step addresses \textbf{RQ1}. We evaluate the effect on intersectional fairness using six metrics described in Section \ref{fairness_definition}: WC-SPD, WC-AOD, WC-EOD, AC-SPD, AC-AOD, and AC-EOD. For each metric, lower values indicate a higher degree of fairness.

Fairness metrics alone can sometimes indicate fairness even if a model is cumulatively biased across all subgroups. To address this, we complement our evaluation with comprehensive ML performance metrics. If a model exhibits cumulative bias (e.g., consistently low performance across all subgroups), its overall performance will be low. Additionally, bias mitigation often comes at the cost of ML performance \cite{sigsoftChenZSH22}. Therefore, it is crucial to also consider the impact on ML performance when evaluating \ours. 
We use five commonly used ML performance metrics: \emph{accuracy}, \emph{precision}, \emph{recall}, \emph{F1-score}, and \emph{MCC} (Matthews Correlation Coefficient). For each metric, higher values correspond to better ML performance.

A brief description of these ML performance metrics is provided as follows. Accuracy indicates the overall correctness of an ML model's predictions. Precision measures the model's accuracy in predicting a specific target class. Recall measures the model's ability to correctly identify all instances of a given target class. F1-score represents the harmonic mean of precision and recall. For precision, recall, and F1-score, we follow previous work \cite{Dabs220703277,sigsoftChenZSH22,ZPICSE24} to employ the macro-average value across favorable and unfavorable classes to comprehensively account for both. This involves computing the metric for each class and then averaging the results. MCC is chosen due to its suitability for imbalanced datasets, which are common in benchmark datasets used for fairness research \cite{Dabs220703277,sigsoftChenZSH22,isstaMoussaS22}. This choice addresses the concern that accuracy, the most widely used metric in fairness research \cite{Dabs220703277}, might not adequately reflect performance in imbalanced class distributions \cite{Dabs220703277,ZPICSE24}.

\subsubsection{Step 4. Comparison of intersectional fairness}\label{incomapre} This step addresses \textbf{RQ2}. We compare the effectiveness of \ours and existing methods in enhancing intersectional fairness. First, we compute the enhancements achieved by each method over the original models and compare these enhancements. 

Additionally, we conduct an in-depth comparison of the fairness metric values obtained by \ours and existing methods through a win-tie-loss analysis \cite{fairsmotepaper,icseLiMC0WZX22}. Specifically, for each existing method, we compare the fairness metric values obtained by \ours and it across the 144 task-fairness metric combinations. Following prior research~\cite{Dabs220703277,sigsoftChenZSH22,ZPICSE24},  we employ the Mann-Whitney U-test \cite{nachar2008mann} to assess the statistical significance of differences in fairness metric values between two methods. The null hypothesis is that there is no difference in fairness metric value distributions between \ours and the existing method, while the alternative hypothesis posits a significant difference. In scenarios where \ours and the existing method exhibit fairness results with statistically significant differences (as indicated by a two-tailed $p$-value~$< 0.05$ from the Mann-Whitney U-test~\cite{Dabs220703277,sigsoftChenZSH22,ZPICSE24}), if \ours yields lower fairness metric values (indicating higher fairness), we label it as a `Win'; if the existing method produces lower fairness metric values, we label \ours as a `Loss.' Conversely, for scenarios where the two methods do not show statistically significant differences, we classify the comparison outcome as a `Tie.'

\subsubsection{Step 5. Comparison  of fairness-performance trade-off}\label{step5}
This step addresses \textbf{RQ3}. 
Bias mitigation often leads to reduced ML performance \cite{sigsoftChenZSH22}. If we assess fairness improvement and ML performance loss separately, it becomes uncertain whether enhanced fairness results from a mere sacrifice in performance \cite{sigsoftHortZSH21}. Moreover, comparing bias mitigation methods while considering these two factors separately can be challenging.
To address this problem, Hort et al. \cite{sigsoftHortZSH21} proposed the use of a unified fairness-performance trade-off baseline, created through their Fairea approach, as a benchmark for comparing different bias mitigation methods. 

To construct the trade-off baseline, Fairea \cite{sigsoftHortZSH21} generates a sequence of mutated models by gradually substituting an increasing portion of the original model's predictions with the majority class prediction from the dataset. This process enhances fairness by uniformly reducing predictive performance across different subgroups. Fairea expects that any reasonable bias mitigation method should outperform these naive mutated models. Therefore, it provides the trade-off attained by these models as the unified baseline for the research community to evaluate bias mitigation methods. 

The trade-off baseline categorizes bias mitigation methods into five levels of trade-off effectiveness \cite{sigsoftHortZSH21}. A method is classified as win-win trade-off if it enhances both ML performance and fairness compared to the original model. Conversely, if a method decreases both, it is categorized as lose-lose trade-off. In cases where a method enhances ML performance but decreases fairness, it is considered inverted trade-off. Additionally, there are two other levels of trade-off where methods decrease ML performance but enhance fairness. Specifically, if a method achieves a superior trade-off compared to the baseline, it is classified as good trade-off; otherwise, it is categorized as poor trade-off. Among all five trade-off types, win-win and good trade-offs indicate that the method surpasses the trade-off baseline constructed by Fairea.

The trade-off analysis covers a total of 30 fairness-performance measurements, since we consider six intersectional fairness metrics and five ML performance metrics.
We use Fairea to construct the trade-off baseline for each pairing of bias mitigation task and fairness-performance measurement. 
We employ these trade-off baselines to comprehensively evaluate the trade-off effectiveness of \ours and existing bias mitigation methods.

\subsubsection{Step 6. Evaluation of mutation strategies}\label{5mutastrategy} This step addresses \textbf{RQ4}. Prior research \cite{icseLiMC0WZX22} has identified correlations between protected attributes and non-protected features, indicating that mutating non-protected features associated with protected attributes could further help improve fairness. Thus, we introduce \ourc, a variant of \ours, for comparison. Unlike \ours, which mutates only protected attributes, \ourc extends mutations to correlated non-protected features. Following prior work \cite{icseLiMC0WZX22}, we use linear regression models to learn these correlations, training models for each non-protected feature using protected attributes as predictors. These models determine the adjustments for non-protected features when protected attributes are mutated.

\subsubsection{Step 7. Evaluation of ensemble strategies} This step addresses \textbf{RQ5}. We compare the three ensemble strategies provided in Section \ref{outputen} in terms of intersectional fairness and fairness-performance trade-off effectiveness.

\subsubsection{Step 8. Analysis of mutant contribution} This step addresses \textbf{RQ6}. We compare the intersectional fairness and fairness-performance trade-off effectiveness of using only mutants involving a single protected attribute and using only mutants involving multiple protected attributes.

\subsubsection{Step 9. Analysis of group fairness} This step addresses \textbf{RQ7}. We evaluate the effect of \ours on group fairness for individual protected attributes. For instance, with the Adult dataset, we assess group fairness regarding sex and race separately. We use SPD, AOD, and EOD as metrics to measure group fairness for each attribute. The Mann-Whitney U-test is also used in this step to ensure statistical significance.

To ensure the reliability of our results, all the experiments in steps 3 to 9 are repeated 20 times.

\section{Results}

\subsection{RQ1: Effect of \ours}
RQ1 investigates the dual effect of \ours on intersectional fairness and ML performance. We apply \ours across 24 tasks, comprising six datasets and four ML models. Given that our experiments are repeated 20 times, we compare the mean metric values over these iterations for both the original models and those applied \ours.

\begin{table*}[!tp]
\begin{threeparttable}
\scriptsize
\centering
\tabcolsep=2.5pt
\caption{(RQ1) Comparative analysis of fairness and ML performance between the original models and \ours. Scenarios where \ours enhances intersectional fairness, indicated by lower fairness metric values, are shaded in grey. \ours improves intersectional fairness in 139 out of 144 task-fairness metric combinations, accounting for 96.5\% of scenarios.}
\label{compa_withdef}
\begin{tabular}{ll|rrrrrrrrrrr|rrrrrrrrrrr}
\hline
\multirow{2}*{Dataset}&&\multicolumn{11}{c|}{LR}&\multicolumn{11}{c}{RF}\\
& & WS & WA & WE & AS & AA & AE & Acc & P & R & F1 & MCC & WS & WA & WE & AS & AA & AE & Acc & P & R & F1 & MCC\\
\hline
\multirow{2}*{Adult}&Original&0.196&0.203&0.322&0.086&0.097&0.157&0.821&0.776&0.683&0.709&0.450&0.213&0.175&0.267&0.089&0.079&0.125&0.839&0.795&0.732&0.754&0.522\\
&\ours&\cellcolor{gray!25}0.101&\cellcolor{gray!25}0.038&\cellcolor{gray!25}0.068&\cellcolor{gray!25}0.041&\cellcolor{gray!25}0.015&\cellcolor{gray!25}0.024&0.812&0.772&0.657&0.682&0.413&\cellcolor{gray!25}0.126&\cellcolor{gray!25}0.044&\cellcolor{gray!25}0.074&\cellcolor{gray!25}0.051&\cellcolor{gray!25}0.016&\cellcolor{gray!25}0.026&0.832&0.792&0.708&0.734&0.493\\
\hline
\multirow{2}*{Compas}&Original&0.507&0.478&0.365&0.187&0.163&0.137&0.674&0.672&0.665&0.666&0.337&0.397&0.368&0.304&0.135&0.120&0.096&0.645&0.641&0.635&0.636&0.276\\
&\ours&\cellcolor{gray!25}0.271&\cellcolor{gray!25}0.255&\cellcolor{gray!25}0.173&\cellcolor{gray!25}0.096&\cellcolor{gray!25}0.085&\cellcolor{gray!25}0.056&0.658&0.676&0.635&0.626&0.308&\cellcolor{gray!25}0.235&\cellcolor{gray!25}0.213&\cellcolor{gray!25}0.175&\cellcolor{gray!25}0.079&\cellcolor{gray!25}0.063&\cellcolor{gray!25}0.053&0.637&0.649&0.614&0.603&0.261\\
\hline
\multirow{2}*{Default}&Original&0.074&0.069&0.103&0.029&0.027&0.039&0.808&0.766&0.599&0.614&0.324&0.083&0.064&0.095&0.034&0.025&0.035&0.813&0.739&0.654&0.677&0.384\\
&\ours&\cellcolor{gray!25}0.059&\cellcolor{gray!25}0.055&\cellcolor{gray!25}0.087&\cellcolor{gray!25}0.025&\cellcolor{gray!25}0.020&\cellcolor{gray!25}0.028&0.811&0.766&0.610&0.628&0.342&0.084&0.066&0.096&\cellcolor{gray!25}0.033&\cellcolor{gray!25}0.024&\cellcolor{gray!25}0.034&0.812&0.734&0.659&0.681&0.386\\
\hline
\multirow{2}*{German}&Original&0.273&0.226&0.232&0.104&0.083&0.088&0.749&0.703&0.669&0.678&0.370&0.201&0.162&0.141&0.070&0.058&0.047&0.758&0.723&0.658&0.670&0.374\\
&\ours&\cellcolor{gray!25}0.150&\cellcolor{gray!25}0.142&\cellcolor{gray!25}0.134&\cellcolor{gray!25}0.051&\cellcolor{gray!25}0.047&\cellcolor{gray!25}0.044&0.745&0.698&0.668&0.676&0.364&\cellcolor{gray!25}0.116&\cellcolor{gray!25}0.109&\cellcolor{gray!25}0.097&\cellcolor{gray!25}0.043&\cellcolor{gray!25}0.038&\cellcolor{gray!25}0.033&0.758&0.728&0.653&0.665&0.373\\
\hline
\multirow{2}*{Mep15}&Original&0.095&0.104&0.168&0.031&0.033&0.054&0.861&0.775&0.663&0.695&0.423&0.077&0.069&0.120&0.024&0.022&0.038&0.860&0.769&0.671&0.702&0.429\\
&\ours&\cellcolor{gray!25}0.059&\cellcolor{gray!25}0.050&\cellcolor{gray!25}0.086&\cellcolor{gray!25}0.021&\cellcolor{gray!25}0.016&\cellcolor{gray!25}0.028&0.859&0.775&0.653&0.686&0.410&\cellcolor{gray!25}0.064&\cellcolor{gray!25}0.052&\cellcolor{gray!25}0.097&\cellcolor{gray!25}0.021&\cellcolor{gray!25}0.016&\cellcolor{gray!25}0.030&0.861&0.769&0.674&0.704&0.432\\
\hline
\multirow{2}*{Mep16}&Original&0.100&0.108&0.172&0.033&0.037&0.061&0.855&0.762&0.641&0.671&0.384&0.066&0.036&0.057&0.021&0.013&0.020&0.855&0.762&0.642&0.672&0.385\\
&\ours&\cellcolor{gray!25}0.042&\cellcolor{gray!25}0.031&\cellcolor{gray!25}0.061&\cellcolor{gray!25}0.014&\cellcolor{gray!25}0.010&\cellcolor{gray!25}0.021&0.854&0.767&0.627&0.657&0.368&\cellcolor{gray!25}0.047&\cellcolor{gray!25}0.030&0.064&\cellcolor{gray!25}0.015&\cellcolor{gray!25}0.010&0.022&0.855&0.763&0.642&0.672&0.386\\
\hline
\multirow{2}*{Dataset}&&\multicolumn{11}{c|}{SVM}&\multicolumn{11}{c}{DNN}\\
& & WS & WA & WE & AS & AA & AE & Acc & P & R & F1 & MCC & WS & WA & WE & AS & AA & AE & Acc & P & R & F1 & MCC \\
\hline
\multirow{2}*{Adult}&Original&0.159&0.138&0.217&0.068&0.063&0.100&0.821&0.788&0.671&0.699&0.444&0.210&0.202&0.316&0.089&0.093&0.149&0.832&0.793&0.709&0.734&0.494\\
&\ours&\cellcolor{gray!25}0.095&\cellcolor{gray!25}0.034&\cellcolor{gray!25}0.061&\cellcolor{gray!25}0.037&\cellcolor{gray!25}0.012&\cellcolor{gray!25}0.020&0.814&0.784&0.652&0.677&0.415&\cellcolor{gray!25}0.111&\cellcolor{gray!25}0.039&\cellcolor{gray!25}0.072&\cellcolor{gray!25}0.044&\cellcolor{gray!25}0.014&\cellcolor{gray!25}0.025&0.821&0.783&0.684&0.708&0.454\\
\hline
\multirow{2}*{Compas}&Original&0.497&0.466&0.353&0.181&0.157&0.130&0.675&0.673&0.665&0.666&0.338&0.495&0.462&0.365&0.180&0.158&0.129&0.673&0.671&0.663&0.663&0.334\\
&\ours&\cellcolor{gray!25}0.287&\cellcolor{gray!25}0.261&\cellcolor{gray!25}0.185&\cellcolor{gray!25}0.099&\cellcolor{gray!25}0.087&\cellcolor{gray!25}0.059&0.661&0.675&0.639&0.632&0.312&\cellcolor{gray!25}0.267&\cellcolor{gray!25}0.256&\cellcolor{gray!25}0.184&\cellcolor{gray!25}0.098&\cellcolor{gray!25}0.085&\cellcolor{gray!25}0.059&0.657&0.674&0.635&0.625&0.306\\
\hline
\multirow{2}*{Default}&Original&0.047&0.046&0.077&0.018&0.017&0.027&0.801&0.769&0.573&0.575&0.279&0.093&0.088&0.134&0.040&0.034&0.047&0.818&0.755&0.656&0.680&0.398\\
&\ours&\cellcolor{gray!25}0.040&\cellcolor{gray!25}0.040&\cellcolor{gray!25}0.073&\cellcolor{gray!25}0.016&\cellcolor{gray!25}0.014&\cellcolor{gray!25}0.024&0.803&0.769&0.580&0.587&0.294&\cellcolor{gray!25}0.087&\cellcolor{gray!25}0.079&\cellcolor{gray!25}0.121&\cellcolor{gray!25}0.037&\cellcolor{gray!25}0.030&\cellcolor{gray!25}0.043&0.816&0.733&0.656&0.675&0.385\\
\hline
\multirow{2}*{German}&Original&0.274&0.232&0.226&0.104&0.084&0.089&0.746&0.699&0.668&0.677&0.366&0.289&0.243&0.242&0.107&0.088&0.096&0.731&0.678&0.651&0.657&0.327\\
&\ours&\cellcolor{gray!25}0.163&\cellcolor{gray!25}0.150&\cellcolor{gray!25}0.165&\cellcolor{gray!25}0.056&\cellcolor{gray!25}0.049&\cellcolor{gray!25}0.052&0.747&0.700&0.672&0.680&0.371&\cellcolor{gray!25}0.172&\cellcolor{gray!25}0.160&\cellcolor{gray!25}0.160&\cellcolor{gray!25}0.061&\cellcolor{gray!25}0.055&\cellcolor{gray!25}0.055&0.740&0.691&0.665&0.670&0.354\\
\hline
\multirow{2}*{Mep15}&Original&0.074&0.069&0.112&0.025&0.022&0.036&0.860&0.776&0.656&0.688&0.415&0.092&0.101&0.166&0.030&0.034&0.058&0.859&0.756&0.660&0.686&0.406\\
&\ours&\cellcolor{gray!25}0.053&\cellcolor{gray!25}0.043&\cellcolor{gray!25}0.074&\cellcolor{gray!25}0.019&\cellcolor{gray!25}0.014&\cellcolor{gray!25}0.024&0.859&0.777&0.651&0.683&0.408&\cellcolor{gray!25}0.057&\cellcolor{gray!25}0.048&\cellcolor{gray!25}0.084&\cellcolor{gray!25}0.020&\cellcolor{gray!25}0.015&\cellcolor{gray!25}0.027&0.857&0.735&0.654&0.676&0.386\\
\hline
\multirow{2}*{Mep16}&Original&0.075&0.061&0.096&0.025&0.022&0.036&0.854&0.764&0.632&0.662&0.373&0.094&0.091&0.143&0.033&0.035&0.056&0.853&0.724&0.642&0.662&0.361\\
&\ours&\cellcolor{gray!25}0.041&\cellcolor{gray!25}0.027&\cellcolor{gray!25}0.051&\cellcolor{gray!25}0.014&\cellcolor{gray!25}0.009&\cellcolor{gray!25}0.019&0.853&0.765&0.622&0.652&0.360&\cellcolor{gray!25}0.047&\cellcolor{gray!25}0.030&\cellcolor{gray!25}0.061&\cellcolor{gray!25}0.015&\cellcolor{gray!25}0.010&\cellcolor{gray!25}0.021&0.853&0.743&0.637&0.661&0.366\\
\hline
\end{tabular}
\begin{tablenotes}
\scriptsize
\item * WS, WA, WE, AS, AA, AE, Acc, P, R, and F1 denote WC-SPD, WC-AOD, WC-EOD, AC-SPD, AC-AOD, AC-EOD, accuracy, precision, recall, and F1-score, respectively.
\end{tablenotes}
\end{threeparttable}
\end{table*}

\begin{table}[!tp]
\scriptsize
\centering
\caption{(RQ1) Mean intersectional fairness and ML performance metric values achieved by the original models and \ours across 24 tasks. \ours enhances intersectional fairness (indicated by decreased fairness metric values) by 40.7\% to 55.3\% across various metrics, while it minimally impacts ML performance, which decreases by only 0.1\% to 2.7\%, depending on the metric analyzed.}
\label{averagechange}
\begin{tabular}{lrr|rr}
\hline
& Original & \ours & \makecell[r]{Absolute\\change} & \makecell[r]{Relative\\change}\\
\hline
WC-SPD & 0.195 & 0.116 & -0.079 & -40.7\%\\
WC-AOD & 0.178 & 0.094 &-0.084&-47.2\%\\
WC-EOD & 0.200&0.104&-0.095&-47.8\%\\
AC-SPD & 0.073 & 0.042 & -0.031 & -42.2\% \\
AC-AOD & 0.065 & 0.031 & -0.034 & -51.8\% \\
AC-EOD & 0.077 & 0.034 & -0.043 & -55.3\%\\
\hline
Accuracy & 0.794&0.791&-0.003&-0.4\%\\
Precision & 0.739&0.738&-0.000&-0.1\%\\
Recall & 0.657&0.648&-0.009&-1.3\%\\
F1-score & 0.675& 0.664&-0.011&-1.6\%\\
MCC & 0.383&0.373&-0.010&-2.7\%\\
\hline
\end{tabular}
\end{table}

Table \ref{compa_withdef} shows the results.
We observe that \ours enhances intersectional fairness, indicated by lower fairness metric values compared to the original models, in 139 out of 144 task-fairness metric combinations, accounting for 96.5\% of scenarios.

Regarding ML performance, we observe a decrease caused by \ours, which can be attributed to the well-recognized fairness-performance trade-off \cite{sigsoftChenZSH22}. However, this decrease in ML performance is significantly outweighed by the substantial improvement in intersectional fairness achieved by \ours. For clarity, we calculate the mean intersectional fairness and ML performance metric values across the 24 tasks for both the original models and \ours. Subsequently, we compute the absolute and relative changes induced by \ours, as depicted in Table~\ref{averagechange}.

Table \ref{averagechange} reveals that \ours enhances intersectional fairness by 40.7\% to 55.3\% (relative changes) across different fairness metrics. In contrast, ML performance experiences a slight decrease ranging from 0.1\% to 2.7\% (relative changes) across various ML performance metrics. 

\finding{\ours largely enhances intersectional fairness at a minimal cost to ML performance. Specifically, it improves intersectional fairness by 40.7\% to 55.3\% across different fairness metrics, while only reducing ML performance by 0.1\% to 2.7\%, depending on the ML performance metric considered.}

\subsection{RQ2: Comparison of Intersectional Fairness}
RQ2 evaluates the intersectional fairness achieved by \ours compared to existing bias mitigation methods. Due to the page limit, statistical results are presented here, omitting detailed fairness metric values for each method and task, which are accessible in our repository \cite{githublink}.

\begin{table*}[!tp]
\scriptsize
\centering
\caption{(RQ2) Mean absolute and relative improvements in intersectional fairness (i.e., decreased fairness metric values) across 24 tasks. We find that \ours consistently outperforms existing bias mitigation methods in enhancing intersectional fairness across all evaluated metrics.}
\label{interfcompare}
\begin{tabular}{lrrrrrrrrrrrr}
\hline
\multirow{2}*{Method}&\multicolumn{2}{c}{WC-SPD} & \multicolumn{2}{c}{WC-AOD} & \multicolumn{2}{c}{WC-EOD} &\multicolumn{2}{c}{AC-SPD} & \multicolumn{2}{c}{AC-AOD} & \multicolumn{2}{c}{AC-EOD} \\
 & Abs. & Rela. & Abs. & Rela. & Abs. & Rela. & Abs. & Rela. & Abs. & Rela. & Abs. & Rela.\\
\hline
REW & -0.025 & -13.0\%&-0.030&-16.7\%&-0.029&-14.4\% & -0.011 & -14.7\% & -0.014 & -21.5\% & -0.017 & -22.0\%\\
ADV & 0.020&10.2\%&0.052&29.4\%&0.060&29.9\% & 0.008 & 10.8\% & 0.018 & 27.6\% & 0.014 & 18.1\%\\
EOP & -0.042&-21.6\%&-0.038&-21.5\%&-0.037&-18.5\% & -0.017 & -22.8\% & -0.016 & -24.8\% & -0.021 & -27.0\%\\
FairSMOTE & -0.024&-12.0\%&-0.032&-18.0\%&-0.038&-19.0\% & -0.013 & -17.4\% & -0.015 & -23.6\%	& -0.022 & -28.0\%\\
MAAT & -0.058&-29.7\%&-0.054&-30.1\%&-0.061&-30.6\% & -0.023 & -31.4\% & -0.022 & -33.6\% & -0.027 & -35.1\%\\
FairMask & -0.055&-28.3\%&-0.070&-39.3\%&-0.073&-36.6\% & -0.023 & -32.1\% & -0.029 & -45.3\% & -0.035 & -45.7\%\\
GRY & -0.069&-35.4\%&-0.062&-34.8\%&-0.048&-23.9\% & -0.026 & -36.4\% & -0.023 & -34.6\% & -0.019 & -24.1\% \\
\ours & -0.079&-40.7\%&-0.084&-47.2\%&-0.095&-47.8\% & -0.031 & -42.2\% & -0.034 & -51.8\% & -0.043 & 	-55.3\%\\
\hline
\end{tabular}
\end{table*}

For each method, we calculate the mean absolute and relative improvements in intersectional fairness across 24 tasks compared to the original models. Results in Table \ref{interfcompare} show that \ours achieves the most substantial enhancement in intersectional fairness, as indicated by the largest decrease in fairness metric values, across all evaluated metrics. Specifically, for WC-SPD, WC-AOD, WC-EOD, AC-SPD, AC-AOD, and AC-EOD, \ours enhances fairness by 40.7\%, 47.2\%, 47.8\%, 42.2\%, 51.8\%, and 55.3\%, respectively.
On average across all metrics, \ours improves intersectional fairness by 47.5\%. In contrast, among existing methods, FairMask achieves the highest improvement in intersectional fairness, with an enhancement of 37.9\% on average across fairness metrics, 9.6 percentage points lower than \ours.

\begin{table}[!tp]
\scriptsize
\centering
\tabcolsep=2.5pt
\caption{(RQ2) Comparative analysis of intersectional fairness between \ours and existing bias mitigation methods across 24 tasks and six fairness metrics. The numbers of \ours's win-tie-loss scenarios are shown. For each metric, \ours consistently outperforms all considered existing methods in terms of scenario wins.}
\label{compare_winlose1}
\begin{tabular}{l|rrrrrrr}
\hline
Metric & REW & ADV & EOP & FairSMOTE & MAAT & FairMask & GRY\\
\hline
WC-SPD & 20/2/2&15/8/1&18/5/1&15/7/2&12/9/3	&15/9/0&13/5/6\\
WC-AOD & 14/9/1&24/0/0&17/7/0&18/3/3&13/8/3&	10/14/0&10/9/5\\
WC-EOD & 14/9/1&24/0/0&17/7/0&11/12/1&	14/7/3&7/17/0&14/9/1\\
AC-SPD & 19/2/3 & 16/5/3 & 16/8/0 & 15/6/3 & 11/10/3 & 16/8/0 & 14/6/4\\
AC-AOD & 14/8/2 & 24/0/0 & 18/6/0 & 17/4/3 & 13/8/3 & 7/17/0 & 10/12/2 \\
AC-EOD & 14/9/1 & 24/0/0 & 18/6/0 & 12/11/1 & 12/9/3& 10/14/0 & 13/10/1\\
\hline
\textbf{Overall} & \textbf{95/39/10} & \textbf{127/13/4} & \textbf{104/39/1} & \textbf{88/43/13}  & \textbf{75/51/18} & \textbf{65/79/0} & \textbf{74/51/19} \\
\hline
\end{tabular}
\end{table}

We further conduct a detailed comparison of fairness metric values obtained by \ours and existing methods using the win-tie-loss analysis outlined in Section \ref{incomapre}. Table \ref{compare_winlose1} shows the results, organizing the win/tie/loss outcomes based on the metrics and also providing an overall summary of the comparison results. From the last row, we find that \ours outperforms all considered existing methods in terms of scenario wins. This indicates that \ours achieves more wins than losses when compared with any of the methods across the six intersectional fairness metrics. For example, \ours surpasses FairMask in 65 scenarios, while FairMask does not outperform \ours in any scenario. Furthermore, when examining each metric individually, we observe that \ours consistently outperforms all considered existing methods.

\finding{\ours consistently outperforms existing bias mitigation methods in enhancing intersectional fairness across all evaluated metrics. On average across all metrics, \ours improves intersectional fairness by 47.5\%, which is 9.6 percentage points higher than that of the currently best-performing method.}

\subsection{RQ3: Comparison of Fairness-Performance Trade-off}
RQ3 employs Fairea \cite{sigsoftHortZSH21}, a state-of-the-art benchmarking tool designed to evaluate fairness-performance trade-off, to compare the trade-off effectiveness of \ours with existing bias mitigation methods. Each method is applied to 24 tasks with 30 fairness-performance measurements, and each experiment is repeated 20 times. As a result, there are a total of $24 \times 30 \times 20 = 14,400$ mitigation cases for each method. Using Fairea, we classify the effectiveness of each method on each case into various effectiveness levels described in Section \ref{step5}, and subsequently, obtain the effectiveness level distributions for each method.

\begin{figure} 
    \centering
\includegraphics[width=1\linewidth]{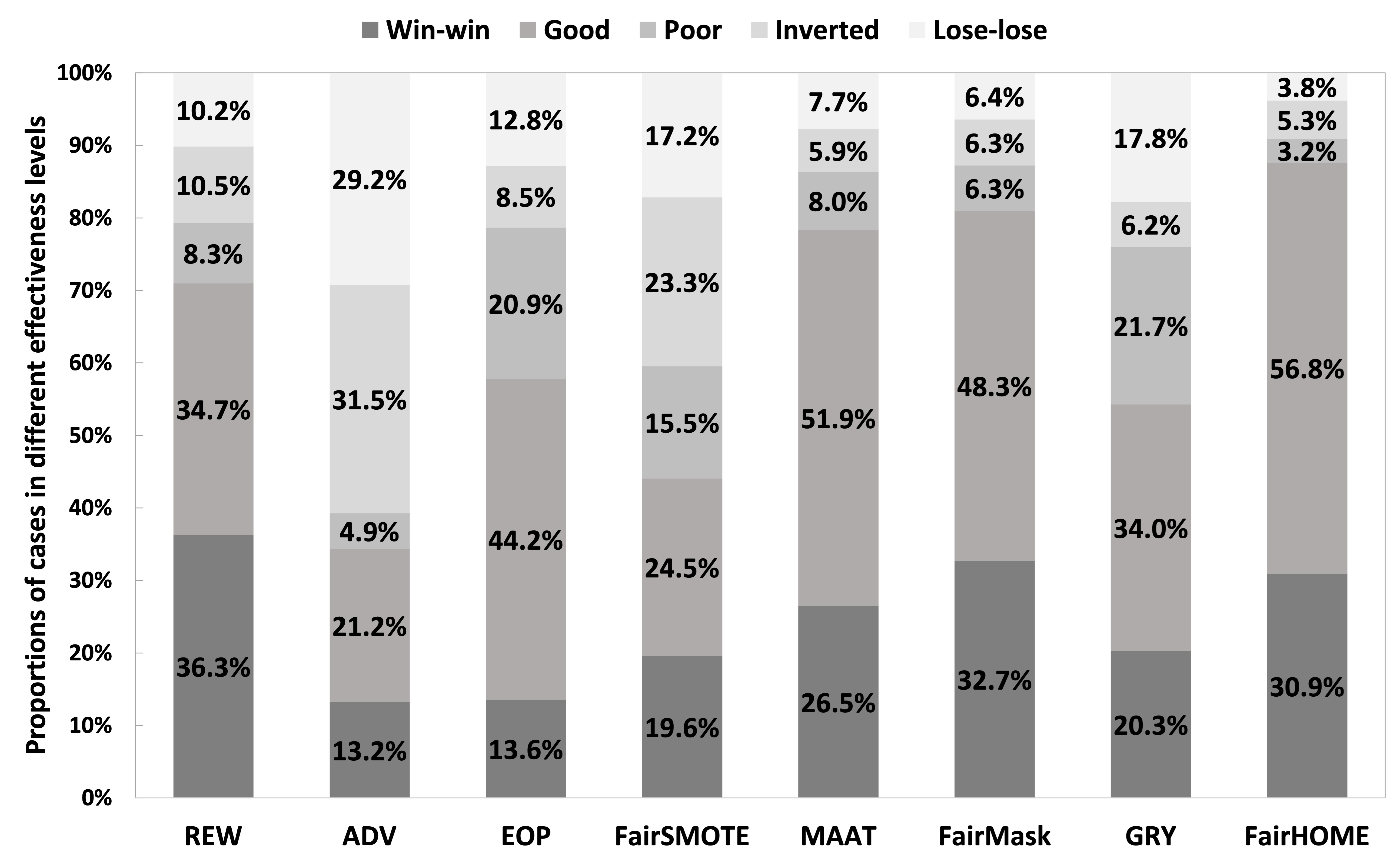}
  \caption{(RQ3) Effectiveness level distributions of \ours and existing methods in fairness-performance trade-off. Overall, \ours achieves the best trade-off, with 87.7\% of mitigation cases falling in win-win or good trade-off.}
  \label{fig:tradeoff} 
\end{figure}

Figure \ref{fig:tradeoff} illustrates the effectiveness level distributions. Overall, \ours demonstrates the best fairness-performance trade-off among all the methods. Specifically, \ours outperforms the trade-off baseline constructed by Fairea (i.e., achieving win-win or good trade-off) in 87.7\% of cases. In contrast, existing methods achieve this in only 34.4\% to 81.0\% of cases. Furthermore, \ours exhibits the fewest mitigation cases falling into the lose-lose trade-off, accounting for 3.8\% of mitigation cases. In contrast, existing methods suffer from lose-lose trade-off in 6.4\% to 29.2\% of cases.

\finding{According to a state-of-the-art benchmarking tool with 30 fairness-performance measurements, \ours outperforms existing bias mitigation methods in fairness-performance trade-off.}

\subsection{RQ4: Evaluation of Mutation Strategies}\label{rq4result}
RQ4 evaluates the impact of mutating not only protected attributes but also features correlated with them within the framework of our approach. We refer to this as \ourc.

\begin{table*}[!tp]
\scriptsize
\centering
\centering
\caption{(RQ4) Mean absolute and relative intersectional fairness improvements achieved by \ours and \ourc across 24 tasks. \ourc achieves lower WC/AC-SPD, but higher WC/AC-AOD and WC/AC-EOD than \ours.}
\label{comparesc}
\begin{tabular}{lrrrrrrrrrrrr}
\hline
\multirow{2}*{Method}&\multicolumn{2}{c}{WC-SPD} & \multicolumn{2}{c}{WC-AOD} & \multicolumn{2}{c}{WC-EOD} &\multicolumn{2}{c}{AC-SPD} & \multicolumn{2}{c}{AC-AOD} & \multicolumn{2}{c}{AC-EOD} \\
 & Abs. & Rela. & Abs. & Rela. & Abs. & Rela. & Abs. & Rela. & Abs. & Rela. & Abs. & Rela.\\
\hline
\ours & -0.079&-40.7\%&-0.084&-47.2\%&-0.095&-47.8\%& -0.031 & -42.2\% & -0.034 & -51.8\% & -0.043 & 	-55.3\%\\
\ourc & -0.120&-61.3\%&-0.081&-45.8\%&-0.082&-41.2\% & -0.048 & -66.1\% & -0.034 & -51.6\% & -0.037 & -48.0\%\\
\hline
\end{tabular}
\end{table*}

First, we evaluate the mean intersectional fairness improvements achieved by \ours and \ourc across 24 tasks. Table \ref{comparesc} presents the results. While \ourc demonstrates improved intersectional fairness through lower WC/AC-SPD values compared to \ours, it also results in higher WC/AC-AOD and WC/AC-EOD values. This suggests that \ourc could exacerbate differences in error rates across subgroups (indicated by increased WC/AC-AOD and WC/AC-EOD), likely due to unintended over-adjustments or noise introduced during the mutation of features correlated with protected attributes.

Second, we compare the effectiveness of the fairness-performance trade-off between \ours and \ourc. For ease of illustration, we follow previous work \cite{sigsoftChenZSH22,ZPICSE24} to measure trade-off effectiveness based on the proportion of mitigation cases where each method surpasses the trade-off baseline constructed by Fairea (i.e., falling in win-win or good trade-off). Overall, \ours surpasses this baseline in 87.7\% of cases, while \ourc does so in 80.4\%, showcasing \ours's superior trade-off between intersectional fairness and performance. This outcome aligns with expectations, as \ourc not only exhibits higher WC/AC-AOD and WC/AC-EOD values, but also tends to introduce additional noise that adversely affects ML performance. 

Considering the drawbacks of \ourc, as evidenced by the results, and its requirement for learning feature correlations, we do not use it as the default strategy of \ours.

\finding{Extending mutations to both protected attributes and their correlated features leads to a poorer fairness-performance trade-off and decreased intersectional fairness across WC/AC-AOD and WC/AC-EOD.}

\subsection{RQ5: Evaluation of Ensemble Strategies}\label{rq5result}
RQ5 evaluates three commonly used ensemble strategies within the framework of our approach: majority vote (the default \ours), averaging (\ourm), and weighted averaging (\oura).

\begin{table*}[!tp]
\scriptsize
\centering
\centering
\caption{(RQ5) Mean absolute and relative intersectional fairness improvements achieved by \ours, \ourm, and \oura across 24 tasks. Overall, they achieve similar intersectional fairness improvements.}
\label{comparesma}
\begin{tabular}{lrrrrrrrrrrrr}
\hline
\multirow{2}*{Method}&\multicolumn{2}{c}{WC-SPD} & \multicolumn{2}{c}{WC-AOD} & \multicolumn{2}{c}{WC-EOD} &\multicolumn{2}{c}{AC-SPD} & \multicolumn{2}{c}{AC-AOD} & \multicolumn{2}{c}{AC-EOD} \\
 & Abs. & Rela. & Abs. & Rela. & Abs. & Rela. & Abs. & Rela. & Abs. & Rela. & Abs. & Rela.\\
\hline
\ours & -0.079&-40.7\%&-0.084&-47.2\%&-0.095&-47.8\% & -0.031 & -42.2\% & -0.034 & -51.8\% & -0.043 & 	-55.3\%\\
\ourm & -0.078&-40.1\%&-0.083&-46.8\%&-0.091&-45.5\% & -0.031 & -42.2\% & -0.034 & -51.5\% & -0.041 & -53.7\%\\
\oura & -0.081&-41.3\%&-0.085&-47.7\%&-0.092&-46.0\% & -0.031 & -43.2\% & -0.034 & -52.2\% & -0.042 & -54.2\%\\
\hline
\end{tabular}
\end{table*}

First, we evaluate the intersectional fairness achieved by the three methods. Table \ref{comparesma} illustrates their mean enhancements across 24 tasks. We find that they achieve similar fairness improvements. Moreover, when compared with the fairness improvements of state-of-the-art methods (as listed in Table \ref{interfcompare}), all three ensemble strategies exhibit superior results.

Second, we evaluate the fairness-performance trade-off effectiveness of the three strategies. We measure this using the proportion of mitigation cases where each method surpasses the trade-off baseline constructed by Fairea (i.e., win-win or good trade-off). We find that \ours, \ourm, and \oura outperform the baseline in 87.7\%, 86.8\%, and 86.6\% of cases, respectively, with a negligible 1.1\% difference. Compared to state-of-the-art methods, which range from 34.4\% to 81.0\% (as shown in Figure \ref{fig:tradeoff}), all three strategies surpass existing methods in fairness-performance trade-off.

In summary, the three ensemble strategies yield comparable results. Given that the majority vote requires only decision information and is simpler, we adhere to the `try-with-simpler' SE practice \cite{sigsoftFuM17} to adopt it as the default ensemble strategy.

\finding{Our approach consistently outperforms existing methods in both intersectional fairness and fairness-performance trade-off under different ensemble strategies: majority vote, averaging, and weighted averaging.}

\subsection{RQ6: Contribution of Different Mutants}
RQ6 compares the effectiveness of three approaches: using only mutants that mutate a single protected attribute (FairHOME4), only mutants that mutate multiple protected attributes (FairHOME5), and a combination of both (\ours). We use the Compas dataset for this RQ, as it includes three protected attributes. For datasets with only two protected attributes, we can generate only one mutant involving mutating multiple attributes, making majority voting impractical for FairHOME5 with only two outputs.

\begin{table*}[!tp]
\scriptsize
\centering
\centering
\caption{(RQ6) Mean absolute and relative intersectional fairness improvements achieved by \ours, FairHOME4, and FairHOME5 across four tasks on the Compas dataset. Overall, \ours performs the best.}
\label{compare45}
\begin{tabular}{lrrrrrrrrrrrr}
\hline
\multirow{2}*{Method}&\multicolumn{2}{c}{WC-SPD} & \multicolumn{2}{c}{WC-AOD} & \multicolumn{2}{c}{WC-EOD} &\multicolumn{2}{c}{AC-SPD} & \multicolumn{2}{c}{AC-AOD} & \multicolumn{2}{c}{AC-EOD} \\
 & Abs. & Rela. & Abs. & Rela. & Abs. & Rela. & Abs. & Rela. & Abs. & Rela. & Abs. & Rela.\\
\hline
\ours & -0.209 & -44.1\% & -0.197 & -44.4\% & -0.168 & -48.4\% & -0.078 & -45.6\% & -0.069 & -46.4\% & -0.066 & 	-53.6\%\\
FairHOME4 & -0.116& -24.4\% &	-0.115& -25.9\%	& -0.113 &	-32.5\% & -0.050 & -29.2\%	& -0.047 & -31.6\%	& -0.044 & -36.0\%\\
FairHOME5 & -0.209 & -44.2\% & -0.185 & -41.8\%	& -0.149 & -42.9\% & -0.080 & -46.9\% & -0.070 &-46.6\% & -0.063 & -51.7\%\\
\hline
\end{tabular}
\end{table*}

Table \ref{compare45} shows the improvements achieved by each model across four tasks on the Compas dataset. FairHOME5 shows greater intersectional fairness improvements compared to FairHOME4. When comparing FairHOME5 to \ours, FairHOME5 achieves an average relative improvement of 45.7\% across six metrics, while \ours achieves 47.3\%. Therefore, \ours performs the best overall. Additionally, we assess the fairness-performance trade-off effectiveness of the three strategies by examining the proportion of cases where each method surpasses the trade-off baseline set by Fairea (i.e., win-win or good trade-off scenarios). The results show that \ours, FairHOME4, and FairHOME5 exceed the baseline in 99.9\%, 96.5\%, and 98.3\% of cases, respectively. This further indicates that using only mutants involving multiple protected attributes outperforms using only mutants involving a single attribute, with the best results achieved by combining both types (i.e., \ours).

\finding{Using mutants involving multiple protected attributes enhances intersectional fairness more than using only single attribute mutations, with the best results achieved by combining both types (i.e., \ours).}

\subsection{RQ7: Effect on Group Fairness}
RQ7 assesses the impact of \ours on group fairness regarding single protected attributes while enhancing intersectional fairness. We evaluate 156 scenarios, combining 13 single-attribute tasks from Table \ref{dataset_info} (e.g., Adult-sex), 4 models, and 3 group fairness metrics (i.e., SPD, AOD, and EOD). According to the Mann-Whitney U-test, \ours significantly improves group fairness regarding single attributes in 120 scenarios and decreases it in 3. In comparison, FairMask, the best baseline approach identified in RQ2 (Table \ref{compare_winlose1}) and RQ3 (Figure \ref{fig:tradeoff}), significantly improves such fairness in 107 scenarios and decreases it in 4.

\ours improves group fairness for single attributes while enhancing intersectional fairness through its comprehensive approach to bias mitigation. By generating mutants representing all possible subgroups, it effectively addresses biases in both single and intersectional attributes. Consequently, \ours's thorough consideration of both single and combined attribute effects leads to simultaneous improvements in group fairness and intersectional fairness.

\finding{In addition to improving intersectional fairness, \ours significantly enhances group fairness for single protected attributes in 120 out of 156 scenarios.}

\section{Discussion}\label{discussion}

\subsection{Advantages of \ours}
\noindent \textbf{Effective.}  The main goal of \ours is to improve intersectional fairness.
Our results in RQ2 indicate that \ours outperforms state-of-the-art methods, resulting in notable enhancements in intersectional fairness.

\noindent \textbf{Balanced.} It is important for bias mitigation methods to strike a balance between fairness and ML performance. 
Our results in RQ3 showcase that \ours outperforms state-of-the-art methods by offering a superior trade-off between intersectional fairness and ML performance.

\noindent \textbf{Non-disruptive.} \ours operates solely on inputs during the inference phase, ensuring seamless implementation without disruption to existing training data processing or necessitating model changes. Even for deployed ML software, applying \ours requires minimal development or deployment efforts, as engineers can easily modify software inputs.

\noindent \textbf{Lightweight access to training data.} Unlike state-of-the-art methods like FairSMOTE, MAAT, and FairMask, which require access to the entire training dataset, \ours accesses only the protected attributes to acquire their possible values, reducing the risk of inadvertently exposing private information within the training data.

\noindent \textbf{No need for training new models.}
In an era where sustainable and green SE practices are increasingly emphasized in both research and industry~\cite{icseGeorgiouK0SZ22}, \ours stands out by eliminating the necessity of training new models.

Moreover, we compare the time cost of different methods. The experiments are executed on Ubuntu 16.04 LTS with 128GB RAM, a 2.3 GHz Intel Xeon E5-2653 v3 Dual CPU, and two NVidia Tesla M40 GPUs. \ours's mutation takes an average of 6.52 seconds across all tasks. In comparison, existing methods that we consider take between 11.48 and 565.20 seconds.

\subsection{Threats to Validity}
\noindent \textbf{Construct validity.} 
The measurement of fairness and ML performance poses a potential threat to the construct validity. To address this concern, we use six intersectional fairness metrics extensively adopted in the literature, alongside five standard ML performance metrics. Additionally, we conduct trade-off analysis using a set of 30 fairness-performance measurements, the most extensive in the fairness literature.

\noindent \textbf{Internal validity.}
To ensure the accuracy of our results, we carefully replicate existing bias mitigation methods for comparative analysis. 
To mitigate the impact of randomness on our results, we conduct 20 repetitions for each experiment. Due to the page limit, we often present average-level statistical results, which may obscure variations and outliers while omitting detailed values for all scenarios. The comprehensive results are available in our repository~\cite{githublink}.

\noindent \textbf{External validity.} 
To address potential concerns regarding external validity, we use 24 bias mitigation tasks for evaluation. These tasks cover six well-studied decision problems and four types of ML models in the fairness literature. When selecting existing methods for comparison, we consider both widely used methods and recent methods. Our selection encompasses pre-processing, in-processing, and post-processing methods.

\section{Conclusion}
This paper introduces \ours, a novel ensemble approach using higher order mutation to improve intersectional fairness of ML software during the inference phase. \ours mutates protected attributes within an input instance to generate diverse inputs from various subgroups. These mutants are then combined with the original input to make the final decision. An extensive evaluation across 24 widely adopted decision tasks demonstrates the power of \ours in surpassing state-of-the-art bias mitigation methods, thereby propelling intersectional fairness to a new height. Moreover, \ours achieves the best trade-off between intersectional fairness and ML performance. 

\section{Data Availability}
We have provided a replication package \cite{githublink}, including all the datasets, code, and intermediate results of our work.

\section*{Acknowledgment}
This research is supported by the National Research Foundation Singapore and DSO National Laboratories under the AI Singapore Programme (AISG Award No: AISG2-RP-2020-019); by the National Research Foundation Singapore and the Cyber Security Agency under the National Cybersecurity R\&D Programme (NCRP25-P04-TAICeN); and by the National Research Foundation, Prime Minister’s Office, Singapore under the Campus for Research Excellence and Technological Enterprise (CREATE) programme. Any opinions, findings, conclusions, or recommendations expressed in this paper are those of the authors and do not reflect the views of the National Research Foundation Singapore or the Cyber Security Agency of Singapore.

\balance
\bibliographystyle{IEEEtran}
\bibliography{fairnessbib}

\begin{thebibliography}{10}
\providecommand{\url}[1]{#1}
\csname url@samestyle\endcsname
\providecommand{\newblock}{\relax}
\providecommand{\bibinfo}[2]{#2}
\providecommand{\BIBentrySTDinterwordspacing}{\spaceskip=0pt\relax}
\providecommand{\BIBentryALTinterwordstretchfactor}{4}
\providecommand{\BIBentryALTinterwordspacing}{\spaceskip=\fontdimen2\font plus
\BIBentryALTinterwordstretchfactor\fontdimen3\font minus \fontdimen4\font\relax}
\providecommand{\BIBforeignlanguage}[2]{{%
\expandafter\ifx\csname l@#1\endcsname\relax
\typeout{** WARNING: IEEEtran.bst: No hyphenation pattern has been}%
\typeout{** loaded for the language `#1'. Using the pattern for}%
\typeout{** the default language instead.}%
\else
\language=\csname l@#1\endcsname
\fi
#2}}
\providecommand{\BIBdecl}{\relax}
\BIBdecl

\bibitem{mahmoud2019performance}
A.~A. Mahmoud, T.~A. Shawabkeh, W.~A. Salameh, and I.~Al~Amro, ``Performance predicting in hiring process and performance appraisals using machine learning,'' in \emph{Proceedings of the 10th international conference on information and communication systems, ICICS 2019}, 2019, pp. 110--115.

\bibitem{donohue2018replacement}
M.~E. Donohue, ``A replacement for justitia's scales: Machine learning's role in sentencing,'' \emph{Harv. JL \& Tech.}, vol.~32, p. 657, 2018.

\bibitem{wu2019investigations}
M.~Wu, Y.~Huang, and J.~Duan, ``Investigations on classification methods for loan application based on machine learning,'' in \emph{Proceedings of the 2019 International Conference on Machine Learning and Cybernetics, ICMLC 2019}, 2019, pp. 1--6.

\bibitem{sigsoftChenZSH22}
Z.~Chen, J.~M. Zhang, F.~Sarro, and M.~Harman, ``{MAAT}: A novel ensemble approach to addressing fairness and performance bugs for machine learning software,'' in \emph{Proceedings of the 30th {ACM} Joint European Software Engineering Conference and Symposium on the Foundations of Software Engineering, {ESEC/FSE} 2022}, 2022, pp. 1122--1134.

\bibitem{icseZhangH21}
J.~M. Zhang and M.~Harman, ``Ignorance and prejudice in software fairness,'' in \emph{Proceedings of the 43rd {IEEE/ACM} International Conference on Software Engineering, {ICSE} 2021}, 2021, pp. 1436--1447.

\bibitem{Dabs220703277}
Z.~Chen, J.~M. Zhang, F.~Sarro, and M.~Harman, ``A comprehensive empirical study of bias mitigation methods for machine learning classifiers,'' \emph{{ACM} Transactions on Software Engineering and Methodology}, vol.~32, no.~4, pp. 106:1--106:30, 2023.

\bibitem{sigsoftBrunM18}
Y.~Brun and A.~Meliou, ``Software fairness,'' in \emph{Proceedings of the 2018 {ACM} Joint Meeting on European Software Engineering Conference and Symposium on the Foundations of Software Engineering, {ESEC/FSE} 2018}, 2018, pp. 754--759.

\bibitem{reHorkoff19}
J.~Horkoff, ``Non-functional requirements for machine learning: Challenges and new directions,'' in \emph{Proceedings of the 27th {IEEE} International Requirements Engineering Conference, {RE} 2019}, 2019, pp. 386--391.

\bibitem{reHabibullahH21}
K.~M. Habibullah and J.~Horkoff, ``Non-functional requirements for machine learning: Understanding current use and challenges in industry,'' in \emph{Proceedings of the 29th {IEEE} International Requirements Engineering Conference, {RE} 2021}, 2021, pp. 13--23.

\bibitem{reHabibullahGH23}
K.~M. Habibullah, G.~Gay, and J.~Horkoff, ``Non-functional requirements for machine learning: Understanding current use and challenges among practitioners,'' \emph{Requir. Eng.}, vol.~28, no.~2, pp. 283--316, 2023.

\bibitem{icseNaharZLK22}
N.~Nahar, S.~Zhou, G.~A. Lewis, and C.~K{\"{a}}stner, ``Collaboration challenges in building {ML}-enabled systems: Communication, documentation, engineering, and process,'' in \emph{Proceedings of the 44th {IEEE/ACM} 44th International Conference on Software Engineering, {ICSE} 2022}, 2022, pp. 413--425.

\bibitem{baresiREnext}
L.~Baresi, C.~Criscuolo, and C.~Ghezzi, ``Understanding fairness requirements for {ML}-based software,'' in \emph{Proceedings of the 31st {IEEE} International Requirements Engineering Conference, {RE} 2023}, 2023, pp. 341--346.

\bibitem{tosemSunCZH24}
Z.~Sun, Z.~Chen, J.~Zhang, and D.~Hao, ``Fairness testing of machine translation systems,'' \emph{{ACM} Transactions on Software Engineering and Methodology}, vol.~33, no.~6, p. 156, 2024.

\bibitem{Dabs220710223}
Z.~Chen, J.~M. Zhang, M.~Hort, M.~Harman, and F.~Sarro, ``Fairness testing: A comprehensive survey and analysis of trends,'' \emph{ACM Transactions on Software Engineering and Methodology}, vol.~33, no.~5, pp. 137:1--137:59, 2024.

\bibitem{fairsmotepaper}
J.~Chakraborty, S.~Majumder, and T.~Menzies, ``Bias in machine learning software: Why? {How}? {What} to do?'' in \emph{Proceedings of the 29th {ACM} Joint European Software Engineering Conference and Symposium on the Foundations of Software Engineering, {ESEC/FSE} 2021}, 2021, pp. 429--440.

\bibitem{fairmaskpaper}
K.~Peng, J.~Chakraborty, and T.~Menzies, ``Fairmask: Better fairness via model-based rebalancing of protected attributes,'' \emph{{IEEE} Transactions on Software Engineering}, vol.~49, no.~4, pp. 2426--2439, 2023.

\bibitem{fairwaypaper}
J.~Chakraborty, S.~Majumder, Z.~Yu, and T.~Menzies, ``Fairway: A way to build fair {ML} software,'' in \emph{Proceedings of the 28th {ACM} Joint European Software Engineering Conference and Symposium on the Foundations of Software Engineering, {ESEC/FSE} 2020}, 2020, pp. 654--665.

\bibitem{icseLiMC0WZX22}
Y.~Li, L.~Meng, L.~Chen, L.~Yu, D.~Wu, Y.~Zhou, and B.~Xu, ``Training data debugging for the fairness of machine learning software,'' in \emph{Proceedings of the 44th {IEEE/ACM} 44th International Conference on Software Engineering, {ICSE} 2022}, 2022, pp. 2215--2227.

\bibitem{sigsoftTaoSHF022}
G.~Tao, W.~Sun, T.~Han, C.~Fang, and X.~Zhang, ``{RULER}: Discriminative and iterative adversarial training for deep neural network fairness,'' in \emph{Proceedings of the 30th {ACM} Joint European Software Engineering Conference and Symposium on the Foundations of Software Engineering, {ESEC/FSE} 2022}, 2022, pp. 1173--1184.

\bibitem{sigsoftNguyenBR23}
G.~Nguyen, S.~Biswas, and H.~Rajan, ``Fix fairness, don't ruin accuracy: Performance aware fairness repair using {AutoML},'' in \emph{Proceedings of the 31st {ACM} Joint European Software Engineering Conference and Symposium on the Foundations of Software Engineering, {ESEC/FSE} 2023}, 2023, pp. 502--514.

\bibitem{icseGoharBR23}
U.~Gohar, S.~Biswas, and H.~Rajan, ``Towards understanding fairness and its composition in ensemble machine learning,'' in \emph{Proceedings of the 45th {IEEE/ACM} International Conference on Software Engineering, {ICSE} 2023}, 2023, pp. 1533--1545.

\bibitem{ZPICSE24}
Z.~Chen, J.~M. Zhang, F.~Sarro, and M.~Harman, ``Fairness improvement with multiple protected attributes: How far are we?'' in \emph{Proceedings of the 46th {ACM/IEEE} International Conference on Software Engineering, {ICSE} 2024}, 2024, pp. 160:1--160:13.

\bibitem{usequal}
``{U.S.} equal employment opportunity commission,'' \url{https://www.eeoc.gov/initiatives/e-race/significant-eeoc-racecolor-casescovering-private-and-federal-sectors\#intersectional}, 2003.

\bibitem{SarroRE23}
F.~Sarro, ``Search-based software engineering in the era of modern software systems,'' in \emph{Proceedings of the 31st IEEE International Requirements Engineering Conferece, RE 2023}, 2023, pp. 3--5.

\bibitem{coleman2017promoting}
P.~T. Coleman, D.~Coon, R.~Kim, C.~Chung, R.~Bass, B.~Regan, and R.~Anderson, ``Promoting constructive multicultural attractors: Fostering unity and fairness from diversity and conflict,'' \emph{The Journal of Applied Behavioral Science}, vol.~53, no.~2, pp. 180--211, 2017.

\bibitem{cimpeanu2023social}
T.~Cimpeanu, A.~Di~Stefano, C.~Perret, and T.~A. Han, ``Social diversity reduces the complexity and cost of fostering fairness,'' \emph{Chaos, Solitons \& Fractals}, vol. 167, p. 113051, 2023.

\bibitem{kim2017diversity}
S.~Kim and S.~Park, ``Diversity management and fairness in public organizations,'' \emph{Public Organization Review}, vol.~17, pp. 179--193, 2017.

\bibitem{zhou2021ensemble}
Z.-H. Zhou, ``Ensemble learning,'' in \emph{Machine learning}.\hskip 1em plus 0.5em minus 0.4em\relax Springer, 2021, pp. 181--210.

\bibitem{sigsoftHortZSH21}
M.~Hort, J.~M. Zhang, F.~Sarro, and M.~Harman, ``Fairea: A model behaviour mutation approach to benchmarking bias mitigation methods,'' in \emph{Proceedings of the 29th {ACM} Joint European Software Engineering Conference and Symposium on the Foundations of Software Engineering, Athens, ESEC/FSE 2021}, 2021, pp. 994--1006.

\bibitem{githublink}
``Replication package,'' \url{https://github.com/chenzhenpeng18/ICSE25-FairHOME}, 2025.

\bibitem{DBcorrabs220707068}
M.~Hort, Z.~Chen, J.~M. Zhang, M.~Harman, and F.~Sarro, ``Bias mitigation for machine learning classifiers: {A} comprehensive survey,'' \emph{ACM Journal on Responsible Computing}, vol.~1, no.~2, pp. 1--52, 2024.

\bibitem{sigsoftGalhotraBM17}
S.~Galhotra, Y.~Brun, and A.~Meliou, ``Fairness testing: Testing software for discrimination,'' in \emph{Proceedings of the 2017 11th Joint Meeting on Foundations of Software Engineering, {ESEC/FSE} 2017}, 2017, pp. 498--510.

\bibitem{biswas2020machine}
S.~Biswas and H.~Rajan, ``Do the machine learning models on a crowd sourced platform exhibit bias? {An} empirical study on model fairness,'' in \emph{Proceedings of the 28th ACM Joint Meeting on European Software Engineering Conference and Symposium on the Foundations of Software Engineering, ESEC/FSE 2020}, 2020, pp. 642--653.

\bibitem{ismail2001use}
R.~Ismail and B.~H. Kleiner, ``The use of the four-fifths rule in discrimination cases,'' \emph{Managerial Law}, vol.~43, no. 1/2, pp. 57--61, 2001.

\bibitem{GhoshGR21}
A.~Ghosh, L.~Genuit, and M.~Reagan, ``Characterizing intersectional group fairness with worst-case comparisons,'' in \emph{Proceedings of the Artificial Intelligence Diversity, Belonging, Equity, and Inclusion, {AIDBEI} 2021}, 2021, pp. 22--34.

\bibitem{emnlpSubramanianHBCF21}
S.~Subramanian, X.~Han, T.~Baldwin, T.~Cohn, and L.~Frermann, ``Evaluating debiasing techniques for intersectional biases,'' in \emph{Proceedings of the 2021 Conference on Empirical Methods in Natural Language Processing, {EMNLP} 2021}, 2021, pp. 2492--2498.

\bibitem{xinyuedriving}
X.~Li, Z.~Chen, J.~M. Zhang, F.~Sarro, Y.~Zhang, and X.~Liu, ``Bias behind the wheel: Fairness testing of autonomous driving systems,'' \emph{ACM Transactions on Software Engineering and Methodology}, 2024.

\bibitem{sigsoftBiswasR21}
S.~Biswas and H.~Rajan, ``Fair preprocessing: Towards understanding compositional fairness of data transformers in machine learning pipeline,'' in \emph{Proceedings of the 29th {ACM} Joint European Software Engineering Conference and Symposium on the Foundations of Software Engineering, {ESEC/FSE} 2021}, 2021, pp. 981--993.

\bibitem{sigsoftZhang022}
M.~Zhang and J.~Sun, ``Adaptive fairness improvement based on causality analysis,'' in \emph{Proceedings of the 30th {ACM} Joint European Software Engineering Conference and Symposium on the Foundations of Software Engineering, {ESEC/FSE} 2022}, 2022, pp. 6--17.

\bibitem{icseGaoZMSCW22}
X.~Gao, J.~Zhai, S.~Ma, C.~Shen, Y.~Chen, and Q.~Wang, ``Fairneuron: Improving deep neural network fairness with adversary games on selective neurons,'' in \emph{Proceedings of the 44th {IEEE/ACM} International Conference on Software Engineering, {ICSE} 2022}, 2022, pp. 921--933.

\bibitem{icseBiswasR23}
S.~Biswas and H.~Rajan, ``Fairify: Fairness verification of neural networks,'' in \emph{Proceedings of the 45th {IEEE/ACM} International Conference on Software Engineering, {ICSE} 2023}, 2023, pp. 1546--1558.

\bibitem{fatBuolamwiniG18}
J.~Buolamwini and T.~Gebru, ``Gender shades: Intersectional accuracy disparities in commercial gender classification,'' in \emph{Proceedings of Conference on Fairness, Accountability and Transparency, {FAT} 2018}, 2018, pp. 77--91.

\bibitem{KangXWMT22}
J.~Kang, T.~Xie, X.~Wu, R.~Maciejewski, and H.~Tong, ``Infofair: Information-theoretic intersectional fairness,'' in \emph{Proceedings of the {IEEE} International Conference on Big Data, Big Data 2022}, 2022, pp. 1455--1464.

\bibitem{fatWangRR22}
A.~Wang, V.~V. Ramaswamy, and O.~Russakovsky, ``Towards intersectionality in machine learning: Including more identities, handling underrepresentation, and performing evaluation,'' in \emph{Proceedings of the 2022 {ACM} Conference on Fairness, Accountability, and Transparency, FAccT 2022}, 2022, pp. 336--349.

\bibitem{icmlKearnsNRW18}
M.~J. Kearns, S.~Neel, A.~Roth, and Z.~S. Wu, ``Preventing fairness gerrymandering: Auditing and learning for subgroup fairness,'' in \emph{Proceedings of the 35th International Conference on Machine Learning, {ICML} 2018}, 2018, pp. 2569--2577.

\bibitem{icseKirschnerSZ20}
L.~Kirschner, E.~O. Soremekun, and A.~Zeller, ``Debugging inputs,'' in \emph{Proceedings of the 42nd International Conference on Software Engineering, {ICSE} 2020}, 2020, pp. 75--86.

\bibitem{infsofJiaH09}
Y.~Jia and M.~Harman, ``Higher order mutation testing,'' \emph{Information Software Technology}, vol.~51, no.~10, pp. 1379--1393, 2009.

\bibitem{sigsoftFuM17}
W.~Fu and T.~Menzies, ``Easy over hard: A case study on deep learning,'' in \emph{Proceedings of the 11th Joint Meeting on Foundations of Software Engineering, {ESEC/FSE} 2017}, 2017, pp. 49--60.

\bibitem{icseZhangW0D0WDD20}
P.~Zhang, J.~Wang, J.~Sun, G.~Dong, X.~Wang, X.~Wang, J.~S. Dong, and T.~Dai, ``White-box fairness testing through adversarial sampling,'' in \emph{Proceedings of the 42nd International Conference on Software Engineering, ICSE 2020}, 2020, pp. 949--960.

\bibitem{tseZhangWSWDWDD22}
P.~Zhang, J.~Wang, J.~Sun, X.~Wang, G.~Dong, X.~Wang, T.~Dai, and J.~S. Dong, ``Automatic fairness testing of neural classifiers through adversarial sampling,'' \emph{{IEEE} Transactions on Software Engineering}, vol.~48, no.~9, pp. 3593--3612, 2022.

\bibitem{ase23fairness}
Z.~Ji, P.~Ma, S.~Wang, and Y.~Li, ``Causality-aided trade-off analysis for machine learning fairness,'' in \emph{Proceedings of the 38th {IEEE/ACM} International Conference on Automated Software Engineering, {ASE} 2023}, 2023, pp. 371--383.

\bibitem{tosemmengdi}
M.~Zhang, J.~Sun, J.~Wang, and B.~Sun, ``{TESTSGD}: Interpretable testing of neural networks against subtle group discrimination,'' \emph{ACM Transactions on Software Engineering and Methodology}, vol.~32, no.~6, pp. 137:1--137:24, 2023.

\bibitem{adultdata}
``Adult census income,'' \url{https://archive.ics.uci.edu/ml/datasets/adult}, 1996.

\bibitem{compasdata}
``Compas,'' \url{https://github.com/propublica/compas-analysis}, 2016.

\bibitem{defaultdata}
``Default,'' \url{https://archive.ics.uci.edu/ml/datasets/default+of+credit+card+clients}, 2016.

\bibitem{germandata}
``German,'' \url{https://archive.ics.uci.edu/ml/datasets/Statlog+\%28German+Credit+Data\%29}, 1994.

\bibitem{mep15data}
``Mep15,'' \url{https://meps.ahrq.gov/mepsweb/data_stats/download_data_files_detail.jsp?cboPufNumber=HC-181}, 2015.

\bibitem{mep16data}
``Mep16,'' \url{https://meps.ahrq.gov/mepsweb/data_stats/download_data_files_detail.jsp?cboPufNumber=HC-192}, 2016.

\bibitem{corrabs220508809}
E.~O. Soremekun, M.~Papadakis, M.~Cordy, and Y.~L. Traon, ``Software fairness: An analysis and survey,'' \emph{CoRR}, vol. abs/2205.08809, 2022.

\bibitem{ukreport}
``Review into bias in algorithmic decision-making,'' \url{https://www.gov.uk/government/publications/cdei-publishes-review-into-bias-in-algorithmic-decision-making/main-report-cdei-review-into-bias-in-algorithmic-decision-making}, 2020.

\bibitem{icseZhengCD0CJW0C22}
H.~Zheng, Z.~Chen, T.~Du, X.~Zhang, Y.~Cheng, S.~Ji, J.~Wang, Y.~Yu, and J.~Chen, ``Neuronfair: Interpretable white-box fairness testing through biased neuron identification,'' in \emph{Proceedings of the 44th {IEEE/ACM} International Conference on Software Engineering, {ICSE} 2022}, 2022, pp. 1519--1531.

\bibitem{rewpaper}
F.~Kamiran and T.~Calders, ``Data preprocessing techniques for classification without discrimination,'' \emph{Knowledge and Information Systems}, vol.~33, no.~1, pp. 1--33, 2011.

\bibitem{ADVpaper}
B.~H. Zhang, B.~Lemoine, and M.~Mitchell, ``Mitigating unwanted biases with adversarial learning,'' in \emph{Proceedings of the 2018 {AAAI/ACM} Conference on AI, Ethics, and Society, {AIES} 2018}, 2018, pp. 335--340.

\bibitem{EOpaper}
M.~Hardt, E.~Price, and N.~Srebro, ``Equality of opportunity in supervised learning,'' in \emph{Proceedings of the Annual Conference on Neural Information Processing Systems 2016, NIPS 2016}, 2016, pp. 3315--3323.

\bibitem{isstaMoussaS22}
R.~Moussa and F.~Sarro, ``On the use of evaluation measures for defect prediction studies,'' in \emph{Proceedings of the 31st {ACM} {SIGSOFT} International Symposium on Software Testing and Analysis, ISSTA 2022}, 2022, pp. 101--113.

\bibitem{nachar2008mann}
N.~Nachar \emph{et~al.}, ``The {Mann-Whitney U}: A test for assessing whether two independent samples come from the same distribution,'' \emph{Tutorials in quantitative Methods for Psychology}, vol.~4, no.~1, pp. 13--20, 2008.

\bibitem{icseGeorgiouK0SZ22}
S.~Georgiou, M.~Kechagia, T.~Sharma, F.~Sarro, and Y.~Zou, ``Green {AI}: Do deep learning frameworks have different costs?'' in \emph{Proceedings of the 44th {IEEE/ACM} International Conference on Software Engineering, {ICSE} 2022}, 2022, pp. 1082--1094.

\end{thebibliography}

\end{document}